\documentclass{article} 
\usepackage{iclr2027_conference,times}

\usepackage{amsmath,amsfonts,bm}

\def\eqref#1{equation~\ref{#1}}

\def\1{\bm{1}}

\DeclareMathAlphabet{\mathsfit}{\encodingdefault}{\sfdefault}{m}{sl}
\SetMathAlphabet{\mathsfit}{bold}{\encodingdefault}{\sfdefault}{bx}{n}

\usepackage{hyperref}
\usepackage{url}

\usepackage{algorithm}
\usepackage{algpseudocode}

\usepackage{amssymb}

\usepackage{graphicx}

\usepackage{wrapfig}

\usepackage{booktabs}

\usepackage[table]{xcolor}

\definecolor{dartgreen}{RGB}{255,242,225}

\usepackage{makecell}

\usepackage{caption}

\usepackage{float}

\title{DART-ES: Difficulty-Aware Reweighting and Targeted Replay for Fine-Tuning LLMs with Evolution Strategies}

\author{%
  Zhishen Sun$^{1}$ \quad
  Hongzhan Wang$^{1}$ \quad
  Sizhe Dang$^{1}$ \quad
  Guang Dai$^{2}$ \quad
  Haishan Ye$^{1,2}$\thanks{Corresponding author.} \\
  $^{1}$Xi'an Jiaotong University \\
  $^{2}$SGIT AI Lab \\
  \texttt{\{szs777,2213722806,darknight1118\}@stu.xjtu.edu.cn} \\
  \texttt{guang.gdai@gmail.com} \\
  \texttt{yehaishan@xjtu.edu.cn}
}

\iclrfinalcopy

\begin{document}

\maketitle
\lhead{Preprint}

\begin{abstract}
Evolution Strategies (ES) enable memory efficient full parameter fine-tuning of large language models (LLMs) using only forward computation. However, standard ES uniformly averages rewards across problems and compresses problem level population feedback into a single scalar, making it difficult to capture how the learning value of each problem changes with model capability. To address this limitation, we propose Difficulty-Aware Reweighting and Targeted Replay for Evolution Strategies (DART-ES). DART-ES estimates the local solvability of each problem from its pass rate across the perturbation population and aggregates historical observations to construct a dynamic difficulty state. This shared state jointly guides continuous difficulty reweighting and rare solvable sample replay, thereby improving perturbation direction evaluation and training data allocation without introducing an additional difficulty model or backpropagation. Extensive experiments show that DART-ES achieves good fine-tuning performance. DART-ES outperforms ES on all five base models and improves the average accuracy from 72.07\% to 73.53\%, exceeding the 73.26\% achieved by GRPO on GSM8K. Across five challenging mathematical reasoning benchmarks, DART-ES achieves an average accuracy of 49.20\%, compared with 48.34\% for ES and remains competitive with strong 7B models trained with RL. Further experiments show consistent gains in instruction tuning, code generation and the Countdown task with a 14B model, demonstrating strong generalization across tasks and scalability to larger models. Beyond performance gains, DART-ES also shows clear advantages in system efficiency. It reduces runtime per step by 15.2\%--50.2\% and peak memory usage per GPU by 21.1\%--51.1\% compared with GRPO. Despite performing full parameter fine-tuning, DART-ES also requires less runtime and GPU memory than GRPO+LoRA. Overall, DART-ES achieves strong performance while maintaining generalization, scalability and system efficiency. Code:\url{https://github.com/szs777/DART-ES-Code}
\end{abstract}

\section{Introduction}
\label{introduction}

Reinforcement Learning with Verifiable Rewards (RLVR) has become an important post-training paradigm for improving the mathematical reasoning ability of large language models (LLMs) \citep{guo2025deepseek, yu2025dapoopensourcellmreinforcement}. Since mathematical problems usually have final answers that can be verified automatically, verifiers based on rules can check the final answer in each model response and assign a scalar reward without requiring an additional learned reward model. RLVR methods such as GRPO use answer correctness to optimize model reasoning behavior and achieve significant progress on a range of mathematical reasoning tasks \citep{shao2024deepseekmath}. However, these methods still rely on backpropagation through policy losses and need to store activations, gradients, and optimizer states for long sequences. Some implementations also maintain additional components such as a reference model, leading to high GPU memory usage and system resource costs that limit their use and scaling in resource constrained settings \citep{sheng2025hybridflow, malladi2023fine}.



Unlike the reinforcement learning (RL) methods, Evolution Strategies (ES) rely only on forward passes. ES directly applies random perturbations to model parameters, evaluates the perturbed models, and optimizes the parameters by aggregating perturbation directions according to population rewards \citep{wierstra2014natural, salimans2017evolution}. It does not need to store training activations, gradients, or optimizer states, resulting in lower GPU memory requirements and easy parallelization. \citet{qiu2025evolution} further show that, with a memory efficient perturbation implementation, ES can perform full parameter fine-tuning of LLMs with billions of parameters without LoRA \citep{hu2021lora} or other dimensionality reduction methods, while achieving performance comparable to RL baselines in several settings. However, standard ES uniformly averages rewards across all problems in a batch. As a result, problems already mastered by the model, problems that the current population cannot solve, and problems solved by only a few perturbation directions receive the same weight during direction evaluation. This treatment ignores their changing learning value and weakens sparse but informative success signals.


To account for the different learning values of problems, recent RLVR studies usually estimate difficulty from multiple rollouts of the same problem \citep{kimiteam2025kimik15scalingreinforcement} and adjust data sampling or update weights accordingly \citep{yu2025dapoopensourcellmreinforcement, bae2026online, zhou2026easy}. However, under binary rewards and small rollout group sizes, group accuracy is sparse and sensitive to sampling noise. As a result, these methods often require more complex problem level weighting schemes \citep{zhang2025grpo, foster2025learning}. In contrast, ES evaluates a relatively large population of parameter perturbed models at each iteration, allowing their results on the same problem to form a more finely resolved population pass rate. This signal captures both the dynamic solvability of the problem relative to the current model and the frequency of potentially effective perturbation directions, providing a natural basis for difficulty aware direction evaluation and training data allocation.

\begin{figure}[t]
  \centering
  \includegraphics[
    page=1,
    width=\linewidth,
    trim={0cm 0.1cm 0cm 0.1cm},
    clip
  ]{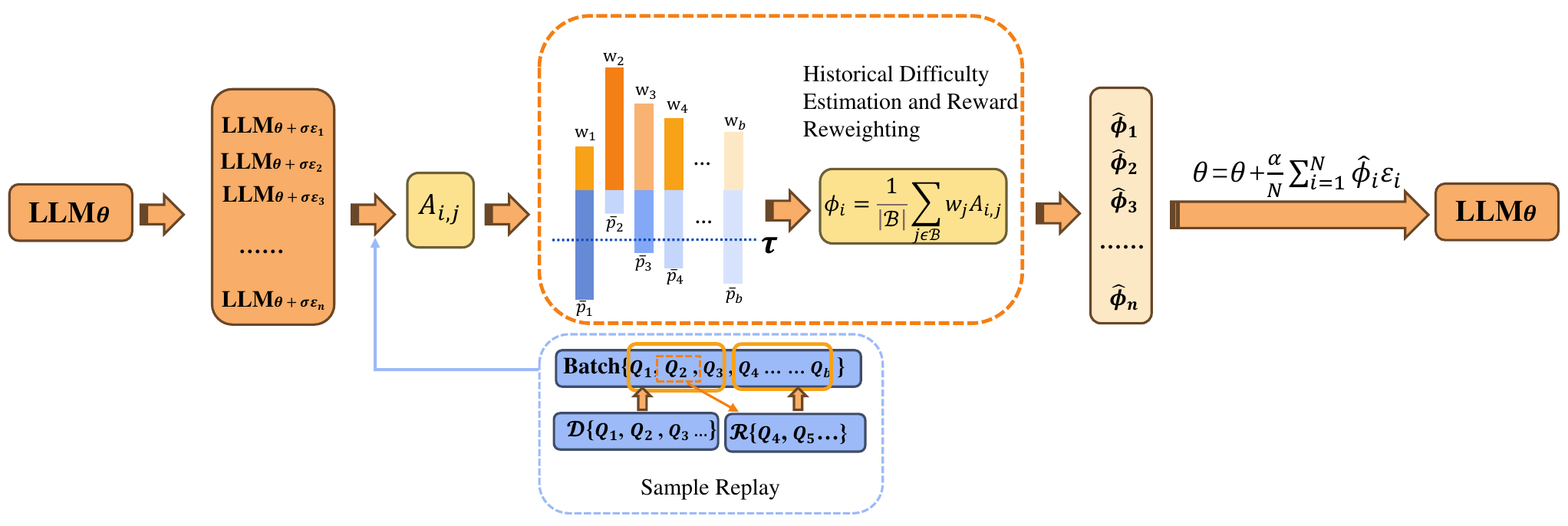}
  \captionsetup{skip=4pt}
  \caption{Overview of the DART-ES workflow.}
  \label{fig:dart-es_flow}
  \vspace{-20pt}
\end{figure}

Based on the above observation, we propose Difficulty-Aware Reweighting and Targeted Replay for Evolution Strategies (DART-ES), a difficulty-aware evolution strategy framework for mathematical reasoning tasks (Figure~\ref{fig:dart-es_flow}). The core idea of DART-ES is to fully use the relatively large perturbation population that ES already evaluates at each iteration. Compared with the sparse accuracy estimates obtained from a small number of rollouts in RL, the performance of different perturbed models on the same problem produces a more informative population pass rate, which describes the solvability of the problem within the current parameter neighborhood in greater detail. Based on this signal, DART-ES aggregates the historical performance of each problem to construct a stable dynamic difficulty state. It then uses difficulty-aware reweighting to increase the influence of high value problems on parameter updates and targeted replay to allocate more training opportunities to difficult but learnable problems. By jointly coordinating perturbation direction evaluation and training data selection, DART-ES improves the use of population information and training resources without introducing an additional difficulty model or backpropagation, while preserving the low memory requirements and easy parallelization of ES.

We conduct extensive experiments to evaluate the effectiveness, generalization and system efficiency of DART-ES. On the relatively simple GSM8K mathematical reasoning benchmark, DART-ES outperforms standard ES on all five base models and achieves the highest average accuracy of 73.53\%, exceeding GRPO at 73.26\%. On five more challenging mathematical reasoning benchmarks, DART-ES reaches an average accuracy of 49.20\% and remains competitive with the strong 7B model trained with RL. We further extend DART-ES to code generation and instruction tuning with a learned reward model and evaluate its scalability on Qwen2.5-14B-It. Across these settings, DART-ES consistently improves standard ES under different tasks, reward forms and model scales. DART-ES also shows clear advantages in runtime and GPU memory usage. It reduces runtime per step by 15.2\%--50.2\% and peak memory per GPU by 21.1\%--51.1\% relative to GRPO, and it also requires less runtime and memory than GRPO+LoRA while performing full parameter fine-tuning. We summarize the contributions of this work as follows:

\begin{itemize}
    \item We propose DART-ES, a difficulty-aware evolution strategy framework for full parameter fine-tuning on mathematical reasoning tasks. It uses problem level correctness information from the perturbation population to define a parameter neighborhood pass rate and aggregates historical observations to construct a dynamic difficulty state.

    \item We design perturbation direction reweighting and rare solvable sample replay mechanisms based on the same historical difficulty state. These mechanisms allow problem level population feedback to guide both current direction evaluation and subsequent data selection, thereby improving the use of population information and training resources.

    \item We conduct extensive experiments across multiple models and tasks. The results show that DART-ES consistently outperforms standard ES, achieves performance comparable to or better than widely used RL methods on mathematical reasoning tasks, and demonstrates strong cross-task generalization and scalability to larger models. In addition, DART-ES achieves shorter runtime per step and lower peak memory usage per GPU than GRPO and GRPO+LoRA, demonstrating good system efficiency.
\end{itemize}

\section{Related Work}
\label{related-work}

\textbf{Reinforcement Learning with Verifiable Rewards.} RLVR \citep{kazemnejad2024vineppo, lambert2024tulu} uses task specific verifiers to evaluate model responses and convert the verification results into scalar rewards. It has been widely applied to automatically verifiable tasks such as mathematics and code. PPO \citep{schulman2017proximal} performs optimization with a clipped policy objective and value estimation, while GRPO \citep{shao2024deepseekmath} estimates advantages using the relative rewards of multiple responses to the same problem and therefore removes the need for a critic model. DeepSeek-R1 \citep{guo2025deepseek} further demonstrates the effectiveness of outcome level verifiable rewards in improving complex reasoning ability.

\textbf{Difficulty Aware Data Selection, Weighting, and Curriculum Learning.} Curriculum learning improves learning efficiency by adjusting the training order or proportion of samples with different difficulty levels \citep{bengio2009curriculum}. This idea is also widely used in LLM reasoning post training, including progressive curriculum learning, offline selection of high value samples, and online difficulty filtering based on the current model performance \citep{parashar2026curriculum, pikus2025hard, li2025limr, yu2025dapoopensourcellmreinforcement, foster2025learning, bae2026online, kong2026rethinking}. In addition, difficulty information is used to adjust policy updates and data reuse, so that difficult but learnable problems have a greater effect during training \citep{zhang2025grpo, dai2026harder, zhou2026easy, sun2026improving}. 

\textbf{Evolution Strategies for Fine Tuning LLMs.} ES perform gradient free optimization by sampling parameter perturbations, evaluating scalar rewards, and aggregating population directions. Natural Evolution Strategies \citep{wierstra2014natural} and OpenAI ES \citep{salimans2017evolution} establish the foundations of this framework from the perspectives of natural gradient optimization and large scale parallelization, respectively. \citet{qiu2025evolution} extend ES to full parameter fine-tuning of billion parameter LLMs and first demonstrate its feasibility for post-training large models. More recently, ESSAM \citep{sun2026essam} and EGGROLL \citep{sarkar2025evolution} further improve ES based LLM fine-tuning from the perspectives of the optimization landscape and perturbation efficiency, respectively.

\section{Method}
\label{method}

In this section, we first introduce the standard ES in Sec.~\ref{jichusuanfa}, and then present the proposed Difficulty-Aware Reweighting and Targeted Replay for Evolution Strategies (DART-ES) in Sec.~\ref{ourssuanfa}.

\subsection{Evolution Strategies}
\label{jichusuanfa}

We adopt the full parameter ES framework introduced by \citet{qiu2025evolution} as our base method. Given an LLM \(\pi\) with initial parameters \(\theta_0\) and a reward function \(r\), our objective is to iteratively optimize the model parameters to maximize the expected reward. At iteration \(t\), the algorithm independently samples \(N\) Gaussian random directions \(\epsilon_i \sim \mathcal{N}(\mathbf{0},\mathbf{I})\), where \(i=1,\ldots,N\), and constructs the corresponding population members as \(\theta_{t,i}=\theta_t+\sigma\epsilon_i\), where \(\sigma\) is the perturbation scale. 

For the current training batch \(\mathcal{B}_t=\{(x_j,y_j)\}_{j=1}^{B_t}\), the average reward of the \(i\)-th perturbed model over the batch is computed as
\begin{equation}
R_i^{(t)}
=
\frac{1}{B_t}
\sum_{j=1}^{B_t}
r\left(
\pi_{\theta_t+\sigma\epsilon_i}(x_j),
y_j
\right).
\end{equation}
After evaluating all population members, we apply z-score normalization and obtain
\(\widehat{R}_i^{(t)}=(R_i^{(t)}-\mu_R^{(t)})/(s_R^{(t)}+\delta)\), where \(\mu_R^{(t)}\) and \(s_R^{(t)}\) denote the mean and standard deviation of the average rewards across the current population, respectively, and \(\delta\) is a small constant for numerical stability. The model parameters are then updated as
\begin{equation}
\theta_{t+1}
=
\theta_t
+
\frac{\alpha}{N}
\sum_{i=1}^{N}
\widehat{R}_i^{(t)}
\epsilon_i.
\label{eq:base_es_update}
\end{equation}
Following \citet{qiu2025evolution}, we absorb the \(1/\sigma\) term into the update scale \(\alpha\) and adopt their memory efficient implementation for full parameter fine-tuning of billion parameter LLMs. The training process requires only forward generation, scalar reward evaluation, and parameter perturbation, without gradients or optimizer states. Further details are provided in Appendix ~\ref{app:es_implementation}.

\begin{algorithm}[!t]
\caption{Difficulty-Aware Reweighting and Targeted Replay for Evolution Strategies}
\label{alg:dart_es}
\footnotesize
\begin{algorithmic}[1]

\Require Initial parameters \(\theta_0\), training set \(\mathcal{D}\),
number of epochs \(E\), population size \(N\), batch size \(B\),
perturbation scale \(\sigma\), update scale \(\alpha\), replay ratio
\(\rho\), EMA decay factor \(\beta\), base weight \(c\), difficulty
exponent \(\kappa\), mastery threshold \(\tau\), zero pass patience
\(K_0\), replay limit \(L\), cooldown interval \(C\), and numerical
stability constant \(\delta\)

\Ensure Fine-tuned model parameters \(\theta\)

\State Initialize \(\theta_t\gets\theta_0\), \(t\gets0\),
replay buffer \(\mathcal{R}\gets\emptyset\),
historical pass rates \(\bar{p}_j\), and zero pass counts \(h_j\gets0\)

\For{\(e=1,\ldots,E\)}
    \State Shuffle \(\mathcal{D}\), initialize the fresh sample cursor,
    and reset the replay counts for the current epoch

    \While{unvisited fresh samples remain}
        \State Build the replay candidate set
        \(\mathcal{C}_t\subseteq\mathcal{R}\) under the mastery,
        replay limit, and cooldown constraints \((\tau,L,C)\)

        \State Construct \(\mathcal{B}_t\) using fresh problems and at most
        \(\operatorname{round}(B\rho)\) replay problems sampled without
        replacement from \(\mathcal{C}_t\), where
        \(\Pr(j\mid\mathcal{C}_t)\propto\exp(-\kappa\bar{p}_j)\)

        \State Update the replay counts and cooldown states of the
        selected replay problems

        \For{\(i=1,\ldots,N\)}
            \State Sample
            \(\epsilon_i\sim\mathcal{N}(\mathbf{0},\mathbf{I})\),
            evaluate \(f_{\theta_t+\sigma\epsilon_i}\) on
            \(\mathcal{B}_t\), and record
            \(A_{i,j}^{(t)}\in\{0,1\}\)
        \EndFor

        \ForAll{\(j\in\mathcal{B}_t\)}
            \State Compute
            \(n_j^{(t)}=\sum_{i=1}^{N}A_{i,j}^{(t)}\) and
            \(p_j^{(t)}=n_j^{(t)}/N\)

            \State Initialize or update
            \(\bar{p}_j\) using EMA decay \(\beta\), update \(h_j\)
            using the zero pass rule, and compute
            \(w_j^{(t)}=c+\exp(-\kappa\bar{p}_j)\)

            \If{\(\bar{p}_j\geq\tau\) or
            \((n_j^{(t)}=0 \text{ and } h_j\geq K_0)\)}
                \State Remove \(j\) from \(\mathcal{R}\)
            \ElsIf{\(n_j^{(t)}>0\)}
                \State Add \(j\) to or keep \(j\) in \(\mathcal{R}\)
            \EndIf
        \EndFor

        \State Compute, for \(i=1,\ldots,N\),
        \[
        \phi_i^{(t)}
        =
        \frac{1}{|\mathcal{B}_t|}
        \sum_{j\in\mathcal{B}_t}
        w_j^{(t)}A_{i,j}^{(t)}
        \]

        \State Normalize
        \(\widehat{\phi}_i^{(t)}
        =(\phi_i^{(t)}-\mu_{\phi}^{(t)})
        /(s_{\phi}^{(t)}+\delta)\)

        \State Update
        \(\theta_{t+1}
        \gets
        \theta_t+
        \frac{\alpha}{N}
        \sum_{i=1}^{N}
        \widehat{\phi}_i^{(t)}\epsilon_i\)

        \State \(t\gets t+1\)
    \EndWhile
\EndFor

\State \Return \(\theta_t\)

\end{algorithmic}
\end{algorithm}

\subsection{Difficulty-Aware Reweighting and Targeted Replay for Evolution Strategies}
\label{ourssuanfa}


Standard ES enables full parameter optimization with low GPU memory usage, but compresses problem level correctness feedback into batch average rewards and assigns equal weights to different problems. This limits its ability to capture the dynamic learning value of each problem relative to the current model and guide subsequent data allocation. To address this limitation, we propose DART-ES, a difficulty-aware evolution strategy framework for mathematical reasoning tasks. While preserving the parameter perturbation and update forms of the ES, DART-ES constructs a dynamic difficulty state from current and historical population pass rates (Sec.~\ref{mokuai1}). This state guides continuous difficulty reweighting (Sec.~\ref{mokuai2}) and rare solvable sample replay (Sec.~\ref{mokuai3}), jointly improving perturbation direction evaluation and training data allocation.
Algorithm~\ref{alg:dart_es} presents the overall procedure. We next introduce the three core components and their implementation details.

\subsubsection{Historical Difficulty Estimation Based on Population Pass Rates}
\label{mokuai1}

At iteration \(t\), we evaluate \(N\) perturbed models on the current training batch
\(\mathcal{B}_t=\{(x_j,y_j)\}_{j=1}^{B_t}\).
Let \(A_{i,j}^{(t)}\in\{0,1\}\) indicate whether the \(i\)-th perturbed model answers the \(j\)-th problem correctly, where \(A_{i,j}^{(t)}=1\) if the final answer is correct and \(A_{i,j}^{(t)}=0\) otherwise. The population evaluation therefore produces a correctness matrix
\(\mathbf{A}^{(t)}\in\{0,1\}^{N\times B_t}\).
For problem \(j\), we define the parameter neighborhood pass rate \(p_j^{(t)}\) as the proportion of perturbed models that answer it correctly. The corresponding number of successful perturbation directions is denoted by \(n_j^{(t)}\):
\begin{equation}
p_j^{(t)}
=
\frac{1}{N}
\sum_{i=1}^{N} A_{i,j}^{(t)},
\qquad
n_j^{(t)}
=
\sum_{i=1}^{N} A_{i,j}^{(t)}
=
Np_j^{(t)}.
\end{equation}
The parameter neighborhood pass rate measures the empirical solvability of a problem within the current sampled parameter neighborhood. A high \(p_j^{(t)}\) indicates that successful directions are common in the current parameter neighborhood. A low but nonzero \(p_j^{(t)}\) indicates that the model cannot yet solve the problem reliably, but a small number of perturbation directions already produce correct answers. When \(p_j^{(t)}=0\), no successful direction is observed in the current population. Therefore, this measure reflects local solvability that changes with the model state, rather than the inherent and fixed difficulty of the problem.

Since a single population evaluation can be affected by random perturbations and the finite population size, we further maintain an exponential moving average of the pass rate for each problem. If problem \(j\) is evaluated for the first time, we initialize its historical state as \(\bar{p}_j^{(t)}=p_j^{(t)}\). For later visits, its historical pass rate is updated as
\begin{equation}
\bar{p}_j^{(t)}
=
\beta\bar{p}_j^{(\mathrm{prev})}
+
(1-\beta)p_j^{(t)},
\end{equation}
where \(\beta\in[0,1)\) is the EMA decay factor, and \(\bar{p}_j^{(\mathrm{prev})}\) denotes the historical state stored after the previous evaluation of problem \(j\). This state is updated only when the problem appears in the current batch. By smoothing the results of multiple population evaluations, \(\bar{p}_j^{(t)}\) reduces short term fluctuations and continuously tracks the dynamic solvability of the problem relative to the current model. It also provides a unified problem level state for subsequent continuous difficulty reweighting and rare solvable sample replay.

\subsubsection{Continuous Difficulty Reweighting Based on Historical Pass Rates}
\label{mokuai2}

Standard ES assigns the same weight to all problems when averaging their rewards within the current batch. To strengthen the learning signals from problems that the model has not yet fully mastered, we assign problem \(j\) a continuous difficulty weight \(w_j^{(t)}=c+\exp(-\kappa\bar{p}_j^{(t)})\), where \(c>0\) is the base weight and \(\kappa>0\) controls how quickly the weight changes with the historical pass rate. Since \(\kappa>0\), \(w_j^{(t)}\) decreases monotonically with \(\bar{p}_j^{(t)}\). Problems with lower historical pass rates therefore receive higher weights, while the base term \(c\) prevents problems that the model can already solve well from being completely ignored.
Based on these weights, the difficulty weighted batch average reward of the \(i\)-th perturbation direction is defined as
\begin{equation}
\phi_i^{(t)}
=
\frac{1}{B_t}
\sum_{j=1}^{B_t}
w_j^{(t)}A_{i,j}^{(t)}.
\label{eq:difficulty_weighted_reward}
\end{equation}
Compared with the batch average reward used in standard ES, \(\phi_i^{(t)}\) depends not only on the number of problems solved by the \(i\)-th perturbed model, but also on their dynamic solvability relative to the current model. When two perturbation directions solve the same number of problems, the direction that solves problems with lower historical pass rates receives a higher batch average reward. This strengthens the learning signals from the small number of successful directions on difficult problems.

When the current population pass rate of problem \(j\) is zero, \(A_{i,j}^{(t)}=0\) for every perturbation direction \(i\). The problem therefore contributes zero to all directions regardless of its historical weight and does not create an artificial direction preference. The reweighting mechanism only strengthens successful directions that are actually observed in the current population.

After computing the difficulty weighted batch average rewards, we normalize them as
\(\widehat{\phi}_i^{(t)}=(\phi_i^{(t)}-\mu_{\phi}^{(t)})/(s_{\phi}^{(t)}+\delta)\),
where \(\mu_{\phi}^{(t)}\) and \(s_{\phi}^{(t)}\) are the mean and standard deviation across the current population, respectively. We then use the normalized difficulty weighted batch average reward \(\widehat{\phi}_i^{(t)}\) to weight the corresponding perturbation direction and update the model parameters according to Eq.~\ref{eq:base_es_update}. Thus, DART-ES changes only the construction of the batch average reward while keeping the parameter perturbation and update form of the base ES unchanged.

\subsubsection{Rare Solvable Sample Replay}
\label{mokuai3}

Continuous difficulty reweighting adjusts the influence of high value problems on the current parameter update, but it cannot determine when these problems are revisited during later training. To address this limitation, we design a rare solvable sample replay mechanism that dynamically maintains a replay buffer using historical pass rates and current population evaluation results. This mechanism provides additional training opportunities for problems that are not yet mastered but already have successful directions in the current parameter neighborhood. To determine whether a problem still has replay value, we maintain a consecutive zero pass count for each problem \(j\):
\begin{equation}
h_j^{(t)}
=
\begin{cases}
h_j^{(\mathrm{prev})}+1, & n_j^{(t)}=0,\\[3pt]
0, & n_j^{(t)}>0.
\end{cases}
\end{equation}
When the current population contains no successful direction, \(h_j^{(t)}\) increases. Once a successful direction appears again, the count is reset to zero. This state complements the historical pass rate. Specifically, \(\bar{p}_j^{(t)}\) describes the long term solvability of the problem, while \(h_j^{(t)}\) identifies problems that continue to lack successful directions.
After completing the current population evaluation and updating these states, the algorithm updates the replay buffer in a fixed order. If \(\bar{p}_j^{(t)}\geq\tau\), problem \(j\) is considered to have reached the mastery threshold and is removed from the buffer. If \(n_j^{(t)}=0\) and \(h_j^{(t)}\geq K_0\), no successful direction appears for \(K_0\) consecutive evaluations, so the problem is also removed. If neither condition is satisfied and \(n_j^{(t)}>0\), the problem is added to or kept in the buffer. In all other cases, its buffer state remains unchanged. Under these rules, the replay buffer mainly retains problems with low historical pass rates that have already shown local solvability. The zero pass patience \(K_0\) allows short term fluctuations and prevents a problem from being removed after only one population evaluation without a successful direction.

Before constructing the current batch, we define the replay priority
of each eligible candidate as
$q_j^{(t)}=\exp(-\kappa\bar p_j^{(\mathrm{prev})})$,
where $\bar p_j^{(\mathrm{prev})}$ denotes the historical state stored
after the most recent evaluation of problem $j$ before the current
iteration. We sample problem $j$ according to
\begin{equation}
\Pr\left(j\mid\mathcal{C}_t\right)
=
\frac{
q_j^{(t)}
}{
\sum_{k\in\mathcal{C}_t}q_k^{(t)}
},
\end{equation}
where \(\mathcal{C}_t\) denotes the set of candidate problems that satisfy all replay constraints. Problems with lower historical pass rates have higher probabilities of receiving additional training opportunities. Replay samples are selected through weighted sampling without replacement. We limit the number of replay visits for each problem within one epoch to \(L\) and set a cooldown interval \(C\) between two replay visits, preventing a small number of difficult problems from being selected too often. The new buffer states and replay priorities produced by the current population evaluation are used only for subsequent batches and do not change the current batch after it has been constructed.

The selected replay samples and fresh samples jointly form the training batch. Let \(B\) denote the maximum batch size and \(\rho\) denote the target replay ratio. The target number of replay samples in each batch is \(B_R=\operatorname{round}(B\rho)\), while the remaining \(B_F=B-B_R\) positions are filled with fresh samples. If there are not enough eligible replay samples, the remaining positions are also filled with fresh samples.



DART-ES uses each problem's historical pass rate to adjust both the batch average rewards of current perturbation directions and the frequency of future problem visits, while continuously updating this state through replay feedback. The method improves the use of population information and computational resources without an additional difficulty prediction model, token level credit assignment, or backpropagation, while preserving the low memory requirements and easy parallelization of the base ES. Further analysis of these mechanisms is provided in Appendix~\ref{app:mechanism_analysis}.

\section{Experiments}
\label{experiment}


We conduct experiments to evaluate the effectiveness and generality of
DART-ES. Section~\ref{gsm8k_ft} reports fine-tuning results on GSM8K,
and Section~\ref{math_ft} examines fine-tuning on MATH and generalization across benchmarks. Section~\ref{Generalization} evaluates generalization and scalability across tasks, reward forms, and model scales through instruction tuning, code generation, and experiments with larger models. Section~\ref{runtime_and_gpu} evaluates the system efficiency of DART-ES by comparing its runtime per step and peak memory usage per GPU with those of the RL baselines. The experimental setup is detailed in Appendix~\ref{Experimental Setup}
and the ablation study is presented in Appendix~\ref{app:ablation_study}.

\begin{table}[H]
\caption{Performance comparison of different fine-tuning methods on GSM8K across base models (\%). Best results are in bold and second-best results are underlined.}
\label{tab:gsm8k-results}
\begin{center}

\small
\setlength{\tabcolsep}{7pt}
\renewcommand{\arraystretch}{1.10}

\begin{tabular}{lcccccc}

\hline
\noalign{\vskip 2pt}

\multicolumn{1}{c}{\bf Method} &
\multicolumn{3}{c}{\bf Qwen2.5-Instruct} &
\multicolumn{2}{c}{\bf LLaMA3.2-Instruct} &
\multicolumn{1}{c}{\bf Avg.}
\\

\cline{2-4}\cline{5-6}
\noalign{\vskip 1pt}

&
\bf 0.5B &
\bf 1.5B &
\bf 3B &
\bf 1B &
\bf 3B &
\\

\noalign{\vskip 2pt}
\hline
\noalign{\vskip 2pt}

Original
& 46.47
& 73.77
& 84.84
& 39.04
& 74.60
& 63.74 \\[1pt]

\hline
\noalign{\vskip 2pt}


GRPO
& \underline{54.24}
& 78.39
& \underline{87.34}
& \textbf{62.32}
& \underline{84.00}
& 73.26 \\

GRPO+LoRA
& 53.30
& 77.94
& 86.13
& 61.41
& 83.32
& 72.42 \\

ES
& 53.44
& 77.25
& 87.11
& 59.66
& 82.87
& 72.07 \\

ESSAM
& 54.06
& \underline{78.92}
& \textbf{87.71}
& \underline{61.79}
& 83.93
& \underline{73.28} \\

\rowcolor{dartgreen}
DART-ES (Ours)
& \textbf{55.65}
& \textbf{79.00}
& \underline{87.34}
& 61.56
& \textbf{84.08}
& \textbf{73.53} \\

\noalign{\vskip 2pt}
\hline

\end{tabular}

\end{center}
\end{table}

\subsection{Fine-Tuning Results on GSM8K}
\label{gsm8k_ft}

Table~\ref{tab:gsm8k-results} presents the fine-tuning results of DART-ES on GSM8K across five base models. In terms of overall performance, DART-ES achieves the highest average accuracy of 73.53\%, showing a consistent improvement over standard ES at 72.07\%. Its average accuracy also exceeds ESSAM at 73.28\% and the RL method GRPO at 73.26\%. It further outperforms the parameter efficient fine-tuning method GRPO+LoRA at 72.42\%. These results show that DART-ES makes more effective use of problem level feedback from population evaluation and incorporates difficulty information into direction evaluation and data allocation. This design further improves the overall fine-tuning performance of standard ES and allows DART-ES to achieve the best average result among the compared methods.

The performance gains of DART-ES remain consistent across different model families and parameter scales. DART-ES outperforms standard ES on all five base models and achieves the best results of 55.65\%, 79.00\% and 84.08\% on Qwen2.5-0.5B-It, Qwen2.5-1.5B-It and LLaMA3.2-3B-It, respectively. DART-ES also matches GRPO with an accuracy of 87.34\% on Qwen2.5-3B-It. Overall, DART-ES reaches or exceeds GRPO on four of the five models, outperforms GRPO+LoRA on every model, and surpasses ESSAM on three models. This consistent trend indicates that the performance improvements of DART-ES are not limited to a particular model architecture or parameter scale.

\subsection{Fine-Tuning Results on MATH}
\label{math_ft}

Table~\ref{tab:reasoning-benchmarks} reports the results of DART-ES on five challenging mathematical reasoning benchmarks. DART-ES achieves an average accuracy of 49.20\%, showing a substantial improvement over the base model Qwen2.5-Math-7B at 26.54\%. Compared with standard ES, DART-ES improves the average accuracy from 48.34\% to 49.20\%. This result shows that difficulty-aware use of population information and sample replay can make better use of informative learning signals from population evaluations and further improve the performance of ES on challenging mathematical tasks. Compared with representative publicly available 7B mathematical reasoning models, DART-ES achieves a higher average score than SimpleRL-Zero-7B and OpenReasoner-Zero-7B. It also remains very close to Oat-Zero-7B which achieves the highest average score, with a gap of 0.44 points. These results show that DART-ES not only substantially improves the reasoning ability of the base model but also achieves strong overall performance on challenging mathematical reasoning tasks.

A closer analysis of the individual benchmarks shows that DART-ES achieves the best results of 35.3\% and 63.9\% on Minerva and AMC, while remaining competitive on other benchmarks. Although the model is trained only on MATH Level 3 to 5 data, it still achieves stable results across multiple mathematical reasoning benchmarks with different problem types and difficulty levels. This indicates that the improvements brought by DART-ES are not limited to a single training benchmark and that the method has good transfer and generalization across benchmarks.

\begin{table}[t]
\caption{Performance comparison of 7B models on challenging math reasoning benchmarks (\%).}
\label{tab:reasoning-benchmarks}
\begin{center}

\small
\setlength{\tabcolsep}{5.5pt}
\renewcommand{\arraystretch}{1.08}

\begin{tabular}{lcccccc}

\hline
\noalign{\vskip 2pt}

{\bf Method} &
{\bf AIME24} &
{\bf Minerva} &
{\bf OlympiadBench} &
{\bf AMC} &
{\bf MATH500} &
{\bf Avg.}
\\

\noalign{\vskip 2pt}
\hline
\noalign{\vskip 2pt}

Qwen2.5-Math-7B
& 10.0
& 8.8
& 17.5
& 43.4
& 53.0
& 26.54 \\

SimpleRL-Zero-7B (GRPO)
& 26.7
& 27.6
& 40.0
& 53.0
& 78.0
& 45.06 \\

OpenReasoner-Zero-7B (PPO)
& 6.7
& 30.1
& \textbf{44.4}
& 42.2
& \textbf{80.6}
& 40.80 \\

Oat-Zero-7B (Dr.GRPO)
& \textbf{33.3}
& 30.5
& \underline{41.9}
& \underline{62.7}
& \underline{79.8}
& \textbf{49.64} \\

ES
& \textbf{33.3}
& \underline{33.1}
& 39.7
& 59.0
& 76.6
& 48.34 \\

\rowcolor{dartgreen}
DART-ES (Ours)
& \underline{30.0}
& \textbf{35.3}
& 39.4
& \textbf{63.9}
& 77.4
& \underline{49.20} \\

\hline

\end{tabular}

\end{center}
\vspace{-5pt}
\end{table}

\begin{table}[t]
\caption{Generalization and scaling performance of DART-ES across tasks and model scales (\%).}
\label{tab:general-code-results}
\begin{center}

\small
\setlength{\tabcolsep}{3.8pt}
\renewcommand{\arraystretch}{1.10}

\begin{tabular}{lcccccc}

\hline
\noalign{\vskip 2pt}

\multicolumn{1}{c}{\bfseries Method} &
\multicolumn{4}{c}{\bfseries Instruction Tuning} &
\multicolumn{1}{c}{\bfseries Code Generation} &
\multicolumn{1}{c}{\bfseries 14B Scaling}
\\

\cline{2-5}\cline{6-6}\cline{7-7}
\noalign{\vskip 1pt}

&
\makecell[c]{\bfseries TruthfulQA} &
\makecell[c]{\bfseries EQ-Bench} &
\makecell[c]{\bfseries MMLU\_Pro} &
\makecell[c]{\bfseries Avg.} &
\makecell[c]{\bfseries LiveCodeBench\\[-1pt]v5} &
\makecell[c]{\bfseries Countdown}
\\

\noalign{\vskip 2pt}
\hline
\noalign{\vskip 3pt}

Original
& 46.54
& 23.86
& 31.86
& 34.09
& 15.05
& 42.2 \\

ES
& \underline{46.58}
& \underline{29.17}
& \underline{32.41}
& \underline{36.05}
& \underline{18.28}
& \underline{63.6} \\

\rowcolor{dartgreen}
DART-ES (Ours)
& \textbf{46.93}
& \textbf{31.59}
& \textbf{32.65}
& \textbf{37.06}
& \textbf{19.71}
& \textbf{67.9} \\

\noalign{\vskip 2pt}
\hline

\end{tabular}

\end{center}
\vspace{-10pt}
\end{table}

\subsection{Generalization and Scalability Performance}
\label{Generalization}

To further examine the generalization and scalability of DART-ES, we extend it to an instruction tuning task with a learned reward model, a code generation task with verifiable rewards, and the larger Qwen2.5-14B-It model. As shown in Table~\ref{tab:general-code-results}, DART-ES outperforms standard ES on all three general alignment and capability benchmarks and achieves the highest average score of 37.06\%, compared with 34.09\% for the base model and 36.05\% for standard ES. The largest improvement appears on EQ-Bench, where DART-ES increases the score from 29.17\% to 31.59\%. It also improves performance on TruthfulQA and MMLU\_Pro. On LiveCodeBench v5, DART-ES achieves an accuracy of 19.71\%, outperforming standard ES at 18.28\%. When scaled to Qwen2.5-14B-It, DART-ES achieves an accuracy of 67.9\% on Countdown, compared with 63.6\% for standard ES. Overall, DART-ES consistently outperforms standard ES in instruction tuning and code generation as well as mathematical reasoning with a 14B model. These results show that its performance gains are not limited to a specific task, reward form or model scale. They also demonstrate strong generalization across tasks and good scalability to larger models.

\begin{figure}[t]
    \centering

    \begin{minipage}[t]{0.49\linewidth}
        \centering
        \includegraphics[width=\linewidth]{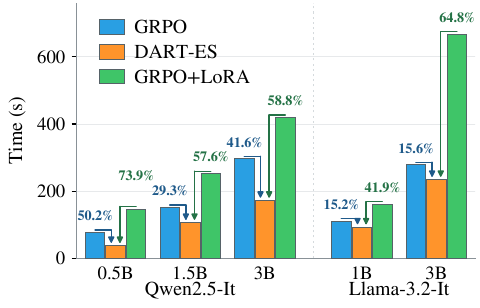}\par
        \vspace{3pt}
        (a) Per-step runtime\par
    \end{minipage}%
    \hfill
    \begin{minipage}[t]{0.49\linewidth}
        \centering
        \includegraphics[width=\linewidth]{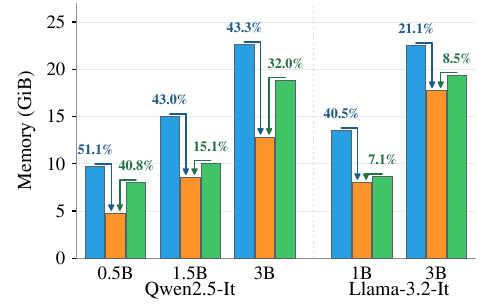}\par
        \vspace{3pt}
        (b) Per-GPU memory\par
    \end{minipage}

    \caption{Runtime and memory comparison across methods and base models: (a) shows the runtime of a single iteration step and (b) shows the mean per-GPU peak memory usage. Detailed experimental settings and results are provided in Appendix ~\ref{Detailed_Per-Step_Runtime_and_GPU_Memory_Results}.}
    \label{efficiency_comparison}
    \vspace{-10pt}
\end{figure}

\subsection{Runtime and GPU Memory Usage}
\label{runtime_and_gpu}

We further compare the mean per step runtime and mean peak memory
usage per GPU of GRPO, GRPO+LoRA, and DART-ES across five base models. As shown in Figure~\ref{efficiency_comparison}, DART-ES achieves the lowest runtime and memory usage in every experimental setting. Compared with GRPO, DART-ES reduces the mean per step runtime by 15.2\%--50.2\% and peak memory usage by 21.1\%--51.1\%. Compared with GRPO+LoRA, it reduces the mean per step runtime by 41.9\%--73.9\% and peak memory usage by 7.1\%--40.8\%. These efficiency gains remain consistent across different model families and parameter scales. Notably, although DART-ES performs full parameter fine-tuning, its per GPU peak memory usage is lower than that of GRPO+LoRA on all five models. In contrast, GRPO+LoRA reduces memory usage relative to GRPO but does not provide a corresponding speed advantage, showing the longest mean per step runtime on every model.

Overall, DART-ES provides consistent advantages in both runtime and memory usage across different model architectures and parameter scales. Compared with GRPO and GRPO+LoRA, DART-ES shortens the per step runtime while further reducing per GPU peak memory, achieving a more favorable balance between computational efficiency and resource usage.

\section{Conclusion}

In this work, we introduced DART-ES, a difficulty-aware evolution strategy framework for full parameter fine-tuning of LLMs. DART-ES constructs a dynamic difficulty state from current and historical population pass rates and uses it to guide continuous difficulty reweighting and rare solvable sample replay. This design jointly improves perturbation direction evaluation and training data allocation without an additional difficulty model or backpropagation. Experiments show that DART-ES consistently improves standard ES on mathematical reasoning and extends effectively to instruction tuning, code generation, and a 14B model. It also reduces runtime per step and peak memory usage per GPU compared with GRPO and GRPO+LoRA under the evaluated settings. These results demonstrate that difficulty-aware population feedback provides an effective and efficient approach to LLMs post-training with evolution strategies.




\bibliography{iclr2027_conference}
\bibliographystyle{iclr2027_conference}

\newpage

\appendix
\section{Detailed ES Procedure and Memory Efficient Implementation}
\label{app:es_implementation}

In this section, we describe the complete training procedure and memory
efficient implementation of ES. Following \citet{qiu2025evolution}, we
reconstruct parameter perturbations from random seeds and use parallel
population evaluation, layerwise in-place perturbation and restoration,
and decomposed in-place updates. These operations avoid storing full
perturbation vectors and aggregated update tensors.

To describe the layerwise implementation, we write
\(\theta_t=\{\theta_{t,\ell}\}_{\ell}\) and
\(\epsilon_i=\{\epsilon_{i,\ell}\}_{\ell}\), where \(\ell\) denotes the
model layer. The complete procedure is shown in
Algorithm~\ref{alg:memory_efficient_es}. In Algorithm~\ref{alg:memory_efficient_es}, $\theta_t$ in the reward expression denotes the
unperturbed reference parameters, while the in-place perturbation
and restoration operations act on each worker's local parameters.

\begin{algorithm}[t]
\caption{Memory Efficient ES for Full Parameter LLM Fine Tuning}
\label{alg:memory_efficient_es}
\footnotesize
\begin{algorithmic}[1]

\Require Initial model parameters \(\theta_0\), training set
\(\mathcal{D}\), per-example reward function \(r\), total iterations
\(T\), population size \(N\), perturbation scale \(\sigma\), update
scale \(\alpha\), number of evaluation workers \(G\), and numerical
stability constant \(\delta\)

\Ensure Fine-tuned model parameters \(\theta_T\)

\State Create \(G\) evaluation workers with identical initial
parameters \(\theta_0\), and designate one worker as the main worker

\For{\(t=0,\ldots,T-1\)}
    \State Sample a training batch
    \(\mathcal{B}_t=\{(x_j,y_j)\}_{j=1}^{B_t}\)
    from \(\mathcal{D}\)

    \State Sample \(N\) random seeds
    \(\{s_i\}_{i=1}^{N}\) and distribute them across the
    \(G\) evaluation workers

    \ForAll{\(i=1,\ldots,N\) \textbf{in parallel}}
        \State Reset the random number generator using \(s_i\)

        \ForAll{model layers \(\ell\)}
            \State Generate
            \(\epsilon_{i,\ell}\sim
            \mathcal{N}(\mathbf{0},\mathbf{I})\)

            \State Perturb the worker-local layer parameters in-place:
            \[
            \theta_{t,\ell}
            \gets
            \theta_{t,\ell}
            +
            \sigma\epsilon_{i,\ell}
            \]
        \EndFor

        \State Compute the batch average reward
        \[
        R_i^{(t)}
        \gets
        \frac{1}{B_t}
        \sum_{j=1}^{B_t}
        r\!\left(
        \pi_{\theta_t+\sigma\epsilon_i}(x_j),
        y_j
        \right)
        \]

        \State Reset the random number generator using \(s_i\)

        \ForAll{model layers \(\ell\)}
            \State Regenerate the same
            \(\epsilon_{i,\ell}\)

            \State Restore the worker-local layer parameters in-place:
            \[
            \theta_{t,\ell}
            \gets
            \theta_{t,\ell}
            -
            \sigma\epsilon_{i,\ell}
            \]
        \EndFor
    \EndFor

    \State Gather the rewards and compute
    \[
    \widehat{R}_i^{(t)}
    \gets
    \frac{
    R_i^{(t)}-\mu_R^{(t)}
    }{
    s_R^{(t)}+\delta
    },
    \qquad
    i=1,\ldots,N
    \]

    \State Initialize the main model parameters as
    \(\theta_{t+1}\gets\theta_t\)

    \For{\(i=1,\ldots,N\)}
        \State Reset the random number generator using \(s_i\)

        \ForAll{model layers \(\ell\)}
            \State Regenerate
            \(\epsilon_{i,\ell}\)

            \State Accumulate the update in-place:
            \[
            \theta_{t+1,\ell}
            \gets
            \theta_{t+1,\ell}
            +
            \frac{\alpha}{N}
            \widehat{R}_i^{(t)}
            \epsilon_{i,\ell}
            \]
        \EndFor
    \EndFor

    \State Synchronize \(\theta_{t+1}\) across all evaluation workers
\EndFor

\State \Return \(\theta_T\)

\end{algorithmic}
\end{algorithm}

\textbf{Noise reconstruction from random seeds.}
A direct implementation of population based ES would store one full
parameter perturbation for every population member, which is
prohibitively expensive for models with billions of parameters. We
instead store only a random seed \(s_i\) for the \(i\)-th population
member. Resetting the random number generator with the same seed
reproduces the same layerwise perturbations
\(\{\epsilon_{i,\ell}\}_{\ell}\) during population evaluation,
parameter restoration, and model update. The implementation therefore
stores only \(N\) random seeds and their scalar rewards rather than
\(N\) full parameter perturbations.

\textbf{Parallel population evaluation.}
The \(N\) perturbed models in each iteration are independent and can be
distributed across \(G\) evaluation workers. Each worker maintains a
local model copy with parameters \(\theta_t\) and processes its assigned
random seeds. After evaluation, the workers return only the batch
average rewards \(R_i^{(t)}\). The main worker performs the parameter
update and then synchronizes \(\theta_{t+1}\) across the remaining
workers. This design allows population evaluation to scale with the
available GPUs and reduces the wall clock time of each iteration.

\textbf{Layerwise in-place perturbation and restoration.}
For seed \(s_i\), the implementation generates the perturbation
\(\epsilon_{i,\ell}\) of each layer and directly adds
\(\sigma\epsilon_{i,\ell}\) to the corresponding parameters
\(\theta_{t,\ell}\). The temporary noise tensor can be released after it
is applied, so the complete perturbation vector \(\epsilon_i\) is never
stored. After forward generation and reward evaluation, the same seed is
used to regenerate the perturbations in the same order. Subtracting
these perturbations restores the worker-local model to \(\theta_t\).
Beyond the model and normal inference states, the perturbation procedure
therefore requires only a temporary noise tensor with the size of one
model layer for each worker.

\textbf{Reward normalization and generation.}
At each iteration, we apply population z-score normalization to the
batch average rewards \(\{R_i^{(t)}\}_{i=1}^{N}\), producing the
normalized rewards \(\{\widehat{R}_i^{(t)}\}_{i=1}^{N}\). This operation
reduces changes in reward scale across iterations and tasks. The
original implementation of \citet{qiu2025evolution} uses greedy decoding
so that differences among population members mainly arise from
parameter space exploration. In our experiments, generation follows the
task-specific settings reported in
Appendix~\ref{Experimental Setup}.

\textbf{Decomposed in-place parameter update.}
Directly constructing
\(\sum_{i=1}^{N}\widehat{R}_i^{(t)}\epsilon_i\)
would require a full aggregated update tensor or multiple full
perturbation vectors. Instead, the main worker reconstructs the
perturbations seed by seed and layer by layer. For the \(i\)-th
population member and layer \(\ell\), the weighted perturbation
\(\widehat{R}_i^{(t)}\epsilon_{i,\ell}\) is directly accumulated into
\(\theta_{t+1,\ell}\) and released before the next layer is processed.
This decomposed update avoids storing a full search direction and
further reduces peak GPU memory usage.

\textbf{Update scale and memory footprint.}
Following \citet{qiu2025evolution}, the \(1/\sigma\) term in the standard
ES update is absorbed into the update scale \(\alpha\). The complete
training procedure does not use backpropagation and does not maintain
gradients or optimizer states. In addition to model parameters and
inference-related states, the implementation stores only random seeds
\(\{s_i\}_{i=1}^{N}\), scalar rewards
\(\{R_i^{(t)}\}_{i=1}^{N}\), and a temporary layerwise noise tensor
\(\epsilon_{i,\ell}\).

Importantly, our proposed DART-ES follows the same random seed based perturbation reconstruction, parameter restoration, and decomposed update pipeline. Its normalized difficulty weighted batch average reward \(\widehat{\phi}_i^{(t)}\) serves as the aggregation weight for perturbation direction \(\epsilon_i\). The difficulty state and sample replay mechanism only modify perturbation direction evaluation and training batch construction, without computing or storing gradients for the full model or maintaining optimizer states.

\clearpage

\section{Experimental Setup}
\label{Experimental Setup}

\textbf{Training and Evaluation Datasets.}
For mathematical reasoning, we train on the GSM8K \citep{cobbe2021training} and MATH \citep{hendrycks2021measuring} datasets. For the relatively simple GSM8K dataset, we fine tune the models on the full training set and evaluate them on the test set. For the more challenging MATH dataset, we fine tune the model on Level 3 to 5 training problems and evaluate its generalization across five widely used mathematical reasoning benchmarks, including OlympiadBench \citep{he2024olympiadbench}, MATH500 \citep{hendrycks2021measuring}, Minerva \citep{lewkowycz2022solving}, AIME2024 \citep{li2024numinamath}, and AMC \citep{li2024numinamath}. For code generation, we train on the Archer-6K dataset \citep{wang2025stabilizing} and evaluate the code generation ability of the model on LiveCodeBench v5 \citep{jain2025livecodebench}. For mathematical reasoning and code generation, we report Pass@1 results. For instruction tuning, we use the UltraFeedback Prompt dataset \citep{cui2023ultrafeedback} for training. After training, we use the \texttt{lm-evaluation-harness} framework \citep{eval-harness} to evaluate the model's general alignment and capabilities on TruthfulQA \citep{lin2022truthfulqa}, EQ-Bench \citep{paech2023eq}, and MMLU\_Pro \citep{wang2024mmlu}, and report the results following the evaluation protocol provided by the repository. For the experiments on larger models, we follow the setup of \citet{qiu2025evolution}. We fine tune the models on the 200 Countdown training examples constructed in their work and evaluate them on the corresponding 2,000 Countdown evaluation examples.


\textbf{Models.}
We select different base models according to the task type and difficulty level. For the relatively simple GSM8K mathematical reasoning task, we use Qwen2.5-0.5B-Instruct, Qwen2.5-1.5B-Instruct, Qwen2.5-3B-Instruct \citep{qwen2.5}, LLaMA3.2-1B-Instruct and LLaMA3.2-3B-Instruct to evaluate DART-ES across different model families and parameter scales. For the more challenging MATH task, we use Qwen2.5-Math-7B \citep{yang2024qwen25mathtechnicalreportmathematical}. For instruction tuning and code generation, we use Qwen2.5-1.5B-Instruct and DeepSeek-R1-Distill-Qwen-1.5B \citep{guo2025deepseek}, respectively. In addition, to further examine the scalability of DART-ES, we conduct experiments on Qwen2.5-14B-Instruct to evaluate its effectiveness on a model with tens of billions of parameters.

\textbf{Baselines.}
On the GSM8K task, we compare DART-ES with GRPO, GRPO+LoRA, standard ES, and ESSAM. GRPO is a representative reinforcement learning method for fine tuning large language models. It estimates advantages from the relative rewards within the response group for each problem and does not require an additional value model. GRPO+LoRA combines GRPO with low rank adaptation to evaluate parameter efficient reinforcement learning with reduced GPU memory usage. Standard ES is a gradient free method for full parameter fine tuning. It updates model parameters by sampling parameter perturbations, evaluating population rewards, and aggregating perturbation directions. The entire process relies only on forward computation and does not store gradients or optimizer states. ESSAM introduces the SAM mechanism into ES and guides parameter search toward flatter high reward regions to improve model generalization. Together, these baselines cover reinforcement learning, parameter efficient reinforcement learning, and different evolution strategy methods, providing representative comparisons for a comprehensive evaluation of DART-ES.

For the more challenging MATH task, we follow the comparison setting of \citet{qiu2025evolution} and compare the model fine tuned with DART-ES against strong publicly available 7B mathematical reasoning models from prior work. These models include SimpleRL-Zero (GRPO) \citep{zeng2025simplerl}, OatZero (Dr.GRPO) \citep{liu2025understanding}, and OpenReasoner (PPO) \citep{hu2026open}, which are trained using well established reinforcement learning frameworks such as VERL \citep{sheng2025hybridflow} and OAT \citep{liu2025oat}. For standard ES, we use the checkpoint with the best average performance reported by \citet{qiu2025evolution}. All results for methods other than DART-ES in Table~\ref{tab:reasoning-benchmarks} are taken from \citet{qiu2025evolution}.

For instruction following with a learned reward model, code generation, and scaling experiments on larger models, we use standard ES as the main fine-tuning baseline to evaluate the performance gains of DART-ES across different tasks, reward forms, and model scales.

\textbf{Hyperparameter settings.}

\textbf{GSM8K Mathematical Reasoning.}
We fine tune all methods on the full GSM8K training set. For DART-ES, the batch size is set to 256 for all base models. We set the base weight to \(c=0.80\), the difficulty exponent to \(\kappa=3.0\), the mastery threshold to \(\tau=0.50\), and the zero pass patience to \(K_0=2\). Each sample is replayed at most once per epoch, with a replay cooldown interval of one iteration. The population size \(N\), perturbation scale \(\sigma\), update scale \(\alpha\), EMA decay factor \(\beta\), and replay ratio \(\rho\) used for different base models are summarized in Table~\ref{tab:gsm8k-hyperparameters}. For each base model, standard ES uses the same population size, perturbation scale, and update scale as DART-ES, while the hyperparameter settings of ESSAM follow those reported in its original work \citep{sun2026essam}. We implement both GRPO and GRPO+LoRA using the VERL framework, with vLLM for rollout generation and Fully Sharded Data Parallel (FSDP) for distributed training. For both methods, the training batch size is set to 256, and eight responses are sampled for each input problem. The maximum prompt length and maximum response length are both set to 512 tokens. The actor mini-batch size is set to 64, with a per-GPU micro-batch size of 8. The rollout model and reference model both use a per-GPU micro-batch size of 16 for log-probability computation. We use a low-variance KL loss with a coefficient of \(0.001\), set the entropy coefficient to 0, and do not include the KL term in the reward. To reduce GPU memory usage, we enable gradient accumulation, padding removal, gradient checkpointing, activation offloading, actor parameter and optimizer state offloading, and reference model parameter offloading. The tensor parallel size for rollout generation is set to 1. The actor learning rate is set to \(1.0\times10^{-6}\) for GRPO and \(1.0\times10^{-5}\) for GRPO+LoRA. For GRPO+LoRA, LoRA is applied to all linear layers, with both the LoRA rank \(r\) and scaling parameter \(\alpha_{\mathrm{LoRA}}\) set to 32.

\begin{table}[t]
\caption{Hyperparameters of DART-ES on GSM8K that vary across base models.}
\label{tab:gsm8k-hyperparameters}
\centering
\small
\setlength{\tabcolsep}{4.0pt}
\renewcommand{\arraystretch}{1.08}

\begin{tabular}{lccccc}
\toprule
\textbf{Base Model}
& \(\boldsymbol{N}\)
& \(\boldsymbol{\sigma}\)
& \(\boldsymbol{\alpha}\)
& \(\boldsymbol{\beta}\)
& \(\boldsymbol{\rho}\) \\
\midrule

Qwen2.5-0.5B-Instruct
& 48
& 0.0012
& \(2.0\times10^{-4}\)
& 0.60
& 0.12 \\

Qwen2.5-1.5B-Instruct
& 64
& 0.0012
& \(2.0\times10^{-4}\)
& 0.60
& 0.15 \\

Qwen2.5-3B-Instruct
& 60
& 0.0015
& \(2.5\times10^{-4}\)
& 0.70
& 0.20 \\

LLaMA3.2-1B-Instruct
& 64
& 0.0012
& \(2.0\times10^{-4}\)
& 0.60
& 0.15 \\

LLaMA3.2-3B-Instruct
& 64
& 0.0015
& \(2.5\times10^{-4}\)
& 0.70
& 0.20 \\

\bottomrule
\end{tabular}
\end{table}

\textbf{MATH Mathematical Reasoning.}
For the more challenging MATH reasoning task, DART-ES uses the following training and generation settings. We set the population size to \(N=30\), the training batch size to 1024, the perturbation scale to \(\sigma=0.001\), and the update scale to \(\alpha=5.0\times10^{-4}\). The maximum generation length is set to 3000 tokens. We set the EMA decay factor to \(\beta=0.70\), the base weight to \(c=0.80\), the difficulty exponent to \(\kappa=3.0\), the replay ratio to \(\rho=0.20\), the mastery threshold to \(\tau=0.50\), and the zero pass patience to \(K_0=2\). Each sample is replayed at most twice per epoch, with a replay cooldown interval of one iteration.

\textbf{Instruction Following with Learned Reward Models.}
For this task, both the base ES and DART-ES run for 100 training
steps to ensure a fair comparison. The two methods use the same main training settings, including a population size of \(N=32\), a training batch size of 256, a perturbation scale of \(\sigma=0.0012\), and an update scale of \(\alpha=2.5\times10^{-4}\). During generation, we use deterministic decoding with temperature set to 0 and top-\(p\) set to 1.0. The maximum prompt length and generation length are set to 768 and 512 tokens, respectively. The base ES updates the model parameters using standard population rewards without introducing an additional difficulty state or sample replay. For DART-ES, we additionally set the EMA decay factor to \(\beta=0.60\), the base weight to \(c=0.80\), the difficulty exponent to \(\kappa=3.0\), the replay ratio to \(\rho=0.15\), the mastery threshold to \(\tau=0.50\), and the zero pass patience to \(K_0=2\). Each sample is replayed at most once per epoch, with a replay cooldown interval of one iteration.

\textbf{Code Generation.}
For the code generation task, the base ES and DART-ES are trained
for 62 iterations using 6,753 training examples from the Archer dataset. Both methods use the same main training settings, including a population size of \(N=30\), a training batch size of 128, a perturbation scale of \(\sigma=0.0012\), and an update scale of \(\alpha=2.0\times10^{-4}\). During generation, the maximum prompt length and maximum generation length are set to 4096 and 32768 tokens, respectively. We set the temperature to 0.6 and top-\(p\) to 0.95. For DART-ES, we further set the EMA decay factor to \(\beta=0.60\), the base weight to \(c=0.80\), the difficulty exponent to \(\kappa=3.0\), the replay ratio to \(\rho=0.15\), the mastery threshold to \(\tau=0.50\), and the zero pass patience to \(K_0=2\). Each sample is replayed at most once per epoch, with a replay cooldown interval of one iteration.

\textbf{Scaling to a Larger Model.}
To further examine the scalability of DART-ES to larger models, we conduct experiments on Qwen2.5-14B-Instruct using the following training settings. For both standard ES and DART-ES, we set the population size to \(N=30\), the training batch size to 200, the perturbation scale to \(\sigma=0.001\), and the update scale to \(\alpha=5.0\times10^{-4}\). The maximum generation length is set to 512 tokens. For DART-ES, we further set the EMA decay factor to \(\beta=0.70\), the base weight to \(c=0.80\), the difficulty exponent to \(\kappa=3.0\), the replay ratio to \(\rho=0.20\), the mastery threshold to \(\tau=0.50\), and the zero pass patience to \(K_0=2\). Each sample is replayed at most twice, with a replay cooldown interval of one iteration.

\textbf{Reward Functions.}
For GSM8K and Countdown, we use a composite reward consisting of an
outcome accuracy reward and an output format reward. Let
\(\hat{y}=\operatorname{extract}(a)\) denote the final answer extracted
from the complete model response \(a\). The accuracy reward is defined as
\begin{equation}
R_{\mathrm{accuracy}}(\hat{y},y)
=
\begin{cases}
1, & \operatorname{is\_equivalent}(\hat{y},y),\\
0, & \text{otherwise}.
\end{cases}
\label{eq:accuracy_reward}
\end{equation}
The format reward is defined as
\begin{equation}
R_{\mathrm{format}}(a)
=
\begin{cases}
1.0, & \text{the output follows the complete required format},\\
0.6, & \text{it contains both markers but does not follow the complete format},\\
0.1, & \text{it contains only the thinking marker},\\
0.5, & \text{it contains only the answer marker},\\
0.0, & \text{otherwise}.
\end{cases}
\label{eq:format_reward}
\end{equation}
The composite reward is
\begin{equation}
R(a,y)
=
R_{\mathrm{accuracy}}(\hat{y},y)
+
0.1R_{\mathrm{format}}(a).
\label{eq:combined_reward}
\end{equation}
The coefficient \(0.1\) keeps the format reward auxiliary to answer
correctness. Equation~\ref{eq:difficulty_weighted_reward} and Algorithm~\ref{alg:dart_es} present the reweighting rule for
binary answer correctness rewards. For GSM8K and Countdown, we retain the
auxiliary format reward and apply difficulty weights only to
the answer correctness component. Let $F_{i,j}^{(t)}$ denote
the format reward of the response generated by the $i$-th
perturbed model for problem $j$. The complete score is
\begin{equation}
    \phi_i^{(t)}
    =
    \frac{1}{B_t}
    \sum_{j=1}^{B_t}
    \left(
        w_j^{(t)} A_{i,j}^{(t)}
        +
        0.1 F_{i,j}^{(t)}
    \right).
    \label{eq:gsm8k_weighted_reward}
\end{equation}
The format reward retains its coefficient of $0.1$ and is not
multiplied by the difficulty weight. These combined scores
are standardized across the population and used to construct
the parameter update. Population pass rates and historical
difficulty states depend only on answer correctness.
Accordingly, the discussion of zero contributions when
$p_j^{(t)}=0$ refers to the answer correctness component.

For MATH, we use only
\(R_{\mathrm{accuracy}}(\hat{y},y)\). For code generation, we use an
execution-based binary reward determined by the test results. 

For the instruction tuning task, we use ArmoRM-Llama3-8B-v0.1
to assign a scalar reward to each model response.
At each iteration, we first evaluate the current unperturbed
model on the same batch to obtain a baseline reward for each
prompt. Let $r_{0,j}^{(t)}$ and $r_{i,j}^{(t)}$ denote the
rewards of the unperturbed model and the $i$-th perturbed model
on prompt $j$, respectively. We define the improvement
indicator and population improvement rate as
\begin{equation}
    I_{i,j}^{(t)}
    =
    \mathbf{1}\!\left[
        r_{i,j}^{(t)} > r_{0,j}^{(t)} + \eta
    \right],
    \qquad
    u_j^{(t)}
    =
    \frac{1}{N}
    \sum_{i=1}^{N} I_{i,j}^{(t)},
    \label{eq:continuous_reward_improvement}
\end{equation}
where $\eta$ is the improvement margin, set to $0$ in our experiments.
We use $u_j^{(t)}$ in place of the population pass rate and
apply the same EMA initialization and update rule to obtain
the historical state $\bar{u}_j^{(t)}$.
The difficulty weight is
$w_j^{(t)} = c + \exp(-\kappa \bar{u}_j^{(t)})$.
Replay follows the same buffer update and sampling rules,
with successful directions defined by the improvement
indicators. The threshold $\tau$ applies to the improvement
rate EMA, and the zero success count tracks consecutive
visits with no direction satisfying the improvement condition.

The score used for the parameter update retains the
continuous rewards,
\begin{equation}
    \phi_i^{(t)}
    =
    \frac{1}{B_t}
    \sum_{j=1}^{B_t}
    w_j^{(t)} r_{i,j}^{(t)}.
    \label{eq:continuous_reward_score}
\end{equation}
We standardize these scores across the population before
constructing the ES update.
Thus, the improvement indicators guide historical state
updates and replay decisions, while direction evaluation
preserves the differences in continuous reward values.

\clearpage
\section{Mechanism Analysis of DART-ES}
\label{app:mechanism_analysis}

\subsection{Effect of Difficulty-Aware Reweighting on Perturbation Direction Evaluation}
\label{app:direction_reweighting}

We conduct experiments on GSM8K with five base models to compare perturbation direction evaluation before and after difficulty-aware reweighting. Specifically, we separately standardize the raw and reweighted rewards for the same population evaluated on the same batch at the same iteration, and compare the corresponding update coefficients. Both scoring rules retain the format reward term, allowing this comparison to show how problem difficulty weights change direction evaluation under the same population feedback.

\begin{figure}[H]
    \centering

    \begin{minipage}[t]{0.32\linewidth}
        \centering
        \includegraphics[width=\linewidth]{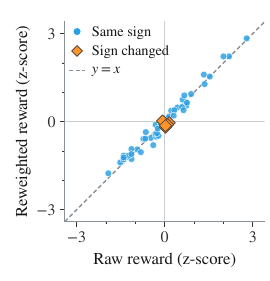}\par
        (a) Qwen2.5-0.5B-It\par
    \end{minipage}%
    \hfill
    \begin{minipage}[t]{0.32\linewidth}
        \centering
        \includegraphics[width=\linewidth]{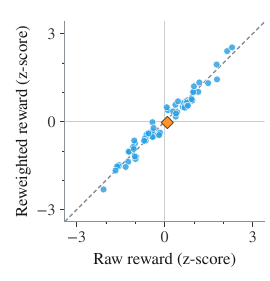}\par
        (b) Qwen2.5-1.5B-It\par
    \end{minipage}%
    \hfill
    \begin{minipage}[t]{0.32\linewidth}
        \centering
        \includegraphics[width=\linewidth]{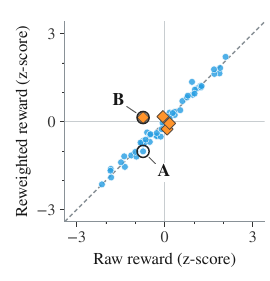}\par
        (c) Qwen2.5-3B-It\par
    \end{minipage}

    \par

    \begin{minipage}[t]{0.32\linewidth}
        \centering
        \includegraphics[width=\linewidth]{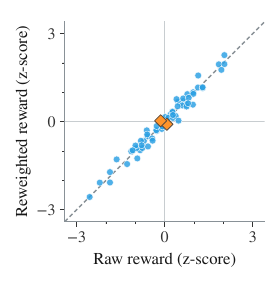}\par
        (d) LLaMA3.2-1B-It\par
    \end{minipage}%
    \hfill
    \begin{minipage}[t]{0.32\linewidth}
        \centering
        \includegraphics[width=\linewidth]{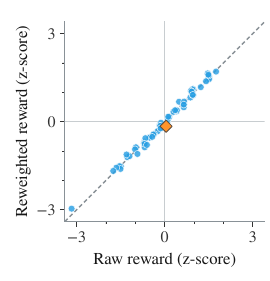}\par
        (e) LLaMA3.2-3B-It\par
    \end{minipage}%
    \hfill
    \begin{minipage}[t]{0.32\linewidth}
        \centering
        \includegraphics[width=\linewidth]{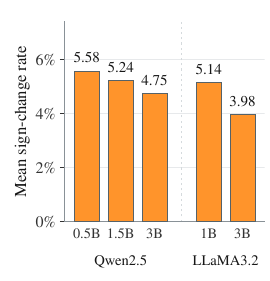}\par
        (f) Sign-change rate\par
    \end{minipage}

    \caption{
    Effect of difficulty-aware reweighting on perturbation coefficients.
    (a--e) Each point compares the separately standardized raw and
    reweighted rewards of the same perturbation in a selected population.
    Blue circles indicate unchanged coefficient signs, orange diamonds
    indicate sign changes, and the dashed line denotes equal coefficients.
    In (c), A and B have identical raw rewards but reweighted coefficients
    of $-1.009$ and $+0.142$, respectively.
    (f) Mean fraction of coefficients that change sign across all recorded
    iterations for each model.
    Comparisons use the same DART-ES populations, and both scoring rules
    retain the format reward term.
}
    \label{fig:direction_reweighting}
    \vspace{-10pt}
\end{figure}

As shown in Figure~\ref{fig:direction_reweighting}, most points lie near the diagonal, indicating a strong correspondence between direction evaluations before and after reweighting. Meanwhile, the standardized coefficients of some directions deviate noticeably from the diagonal, with some changing from negative to positive or from positive to negative. The fraction of coefficients that change sign, averaged over iterations for each model, ranges from 3.98\% to 5.58\% across the five base models. This shows that problem difficulty information directly affects the update coefficients. A more direct example comes from one population evaluation of Qwen2.5-3B-It. The two perturbation directions marked A and B have the same answer accuracy and raw reward, so the unweighted scoring rule assigns them the same standardized coefficient. After problem difficulty weights are introduced, their coefficients become $-1.009$ and $+0.142$, respectively. This difference arises because the sums of the weights assigned to correctly solved problems differ between the two directions. Thus, even when two directions solve the same number of problems correctly, reweighting distinguishes them according to the historical solvability of those problems. Differences in solution outcomes at the problem level therefore enter direction evaluation and affect the corresponding update coefficients.

These results highlight the design advantage of difficulty-aware reweighting. DART-ES incorporates both current solution outcomes and historical problem solvability into direction evaluation, assigning different update coefficients to perturbation directions that raw average rewards cannot distinguish. By using historical information at the problem level to differentiate the contributions of perturbation directions, DART-ES makes fuller use of the information about which problems each direction solves. Parameter updates therefore account for both the number of successful solutions and the difficulty of the corresponding problems.

\subsection{Effect of Targeted Replay on Training Sample Allocation}
\label{app:replay_sample_allocation}

We compare the distributions of current population pass rates for regular traversal and replay visits across five base models on GSM8K. Here, $p$ denotes the fraction of models in the current perturbation population that correctly solve a given problem. Within each epoch, we calculate the fraction of visits in each pass rate interval separately for each sampling source, using the actual number of visits from that source as the denominator. Repeated visits to the same problem are counted separately, so the distribution reflects the allocation of training opportunities across problems with different levels of solvability.

\begin{figure}[t]
    \centering
    \includegraphics[
        page=1,
        width=\linewidth
    ]{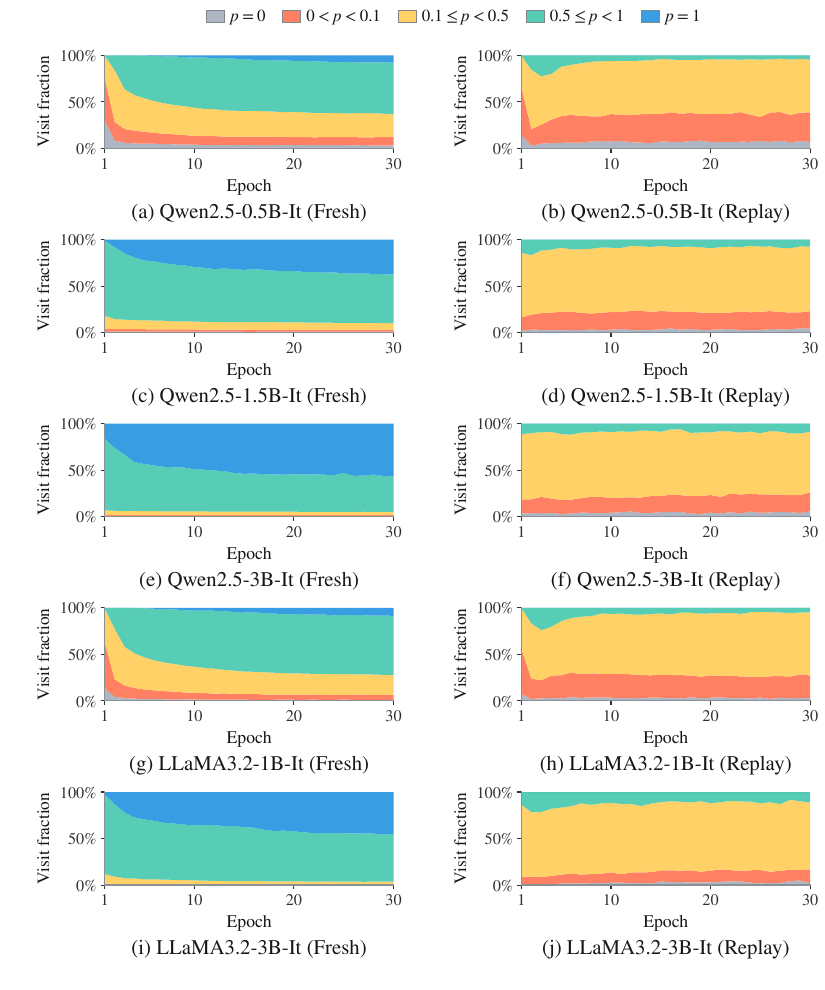}
    \captionsetup{skip=4pt}
    \caption{
    Population solvability during DART-ES training on GSM8K.
    Each row corresponds to a base model.
    The left and right columns show regular traversal (fresh)
    and replay visits, respectively.
    Stacked areas show the fraction of visits in each current
    population pass rate interval, computed separately for each
    epoch and sampling source.
    Here, $p$ is the fraction of perturbed models that solve
    a problem correctly.
    The intervals describe current population observations
    and do not denote mastery states based on historical pass rates.
    }
    \vspace{-10pt}
    \label{fig:replay_solvability}
\end{figure}

As shown in Figure~\ref{fig:replay_solvability}, the fraction of regular traversal visits to problems with higher pass rates generally increases as training progresses, while replay visits in DART-ES remain concentrated on problems with low but nonzero pass rates. Aggregating actual visit counts, we find that the fraction of replay visits with $0<p<0.5$ ranges from 83.94\% to 88.95\% across the five base models, compared with 4.45\% to 40.97\% for regular traversal. For Qwen2.5-3B-It, for example, this interval accounts for 4.76\% of regular traversal visits and 86.61\% of replay visits. This difference reflects how targeted replay adjusts the allocation of training opportunities. Even as regular traversal visits increasingly involve easier problems, replay continues to direct additional visits toward problems that the current population does not yet solve reliably, maintaining training opportunities for these problems.

This visit distribution reflects the replay mechanism's focus on both problem difficulty and local solvability. When $p=0$ or $p=1$, correctness feedback for a given problem is identical across the current population and cannot distinguish perturbation directions based on answer correctness for that problem. For a problem with $0<p<0.5$, a minority of population members solve it successfully, while most members still fail. This indicates that the problem remains challenging for the current population and that observable successful directions already exist in the current parameter neighborhood. Such problems therefore provide feedback at the problem level that includes both successes and failures, allowing perturbation direction evaluation to use concrete differences in solution outcomes. The high fraction of replay visits to these problems shows that additional training opportunities mainly go to problems that remain challenging and offer observed successes as a reference. This matches the method's design goal of using success signals from difficult problems.

These results show that DART-ES consistently directs additional training opportunities toward problems that remain challenging for the current population but already have successful directions. This highlights the advantage of a targeted replay design that considers both difficulty and local solvability. By using the historical difficulty state to guide sample selection, targeted replay makes additional visits more focused and complements difficulty-aware reweighting. Together, these mechanisms use population feedback at the problem level to guide both direction evaluation and the allocation of training opportunities.

\subsection{Smoothing and Dynamic Tracking of Historical Difficulty States}
\label{app:historical_difficulty_tracking}

DART-ES uses an exponential moving average (EMA) to integrate population feedback at the problem level and uses the same historical difficulty state to guide perturbation direction evaluation and replay selection. We examine the role of historical information through predictions of subsequent pass rates and state trajectories for individual problems. Specifically, we use the population pass rate from the preceding visit to a problem and the EMA updated at that visit as separate predictors of the pass rate at the next actual visit to the same problem. The population pass rate is the fraction of members in the current perturbation population that solve the problem correctly. Both predictors use only information available before the target visit, and we compute their mean squared error (MSE) on the same target visits. Results are grouped by model, target visit source, and training stage. Relative MSE is the prediction MSE of EMA divided by that of the preceding pass rate, with values below 1 indicating lower error for EMA.

\begin{figure}[H]
    \centering
    \includegraphics[
        page=1,
        width=\linewidth
    ]{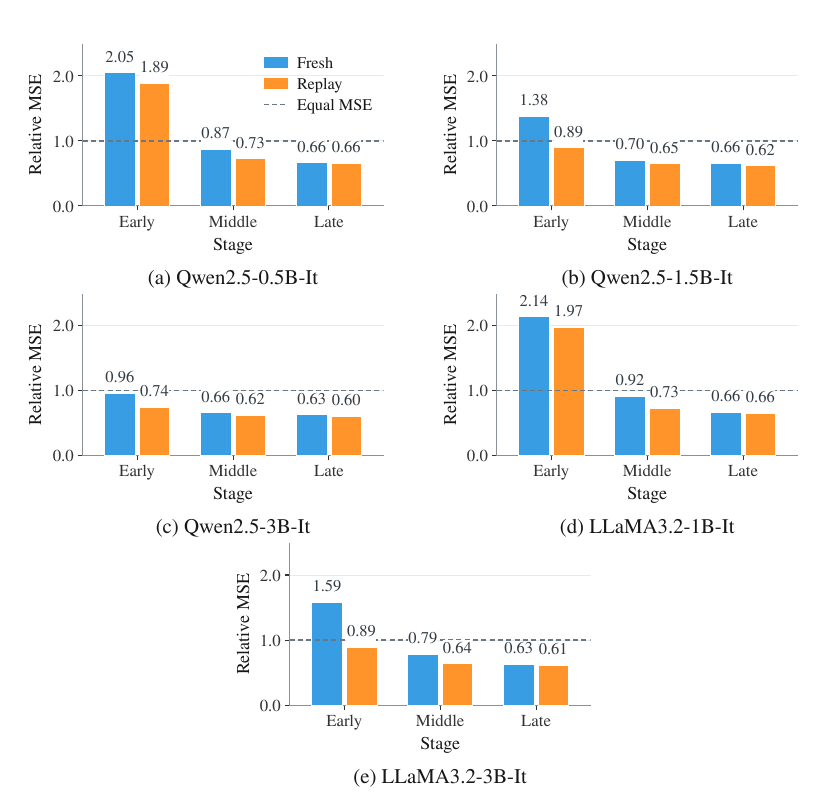}
    \captionsetup{skip=4pt}
    \caption{
    Population pass rate prediction error during DART-ES training on GSM8K.
    Each panel corresponds to a base model.
    Blue and orange bars represent regular traversal and replay target
    visits, respectively.
    Relative MSE is the prediction MSE of the preceding visit's updated
    EMA divided by that of its pass rate, with both predicting the next
    observed pass rate for the same problem.
    Values below 1 indicate lower error for EMA.
    Early and Late correspond to the first and last five complete epochs,
    respectively; Middle contains the remaining complete epochs.
}
    \label{fig:ema_prediction}
    \vspace{-15pt}
\end{figure}

Accumulated historical information provides a consistent predictive advantage in the middle and late stages of training. As shown in Figure~\ref{fig:ema_prediction}, relative MSE is below 1 for both regular traversal and replay visits across all five base models in these two stages. MSE decreases by 8.46\% to 38.10\% in the middle stage and by 33.51\% to 39.55\% in the late stage. However, relative MSE exceeds 1 for some models in the early stage, which is consistent with a possible response lag from historical smoothing when solvability changes rapidly. After combining both types of target visits from all complete epochs within each model, EMA has lower overall error for three models. For Qwen2.5-0.5B-It and LLaMA3.2-1B-It, its overall MSE is 22.08\% and 29.27\% higher than using only the most recent pass rate, respectively. Thus, EMA uses accumulated feedback more effectively in the middle and late stages, while also reflecting a tradeoff between smoothing and response speed.

\begin{figure}[t]
    \centering
    \includegraphics[
        page=1,
        width=\linewidth
    ]{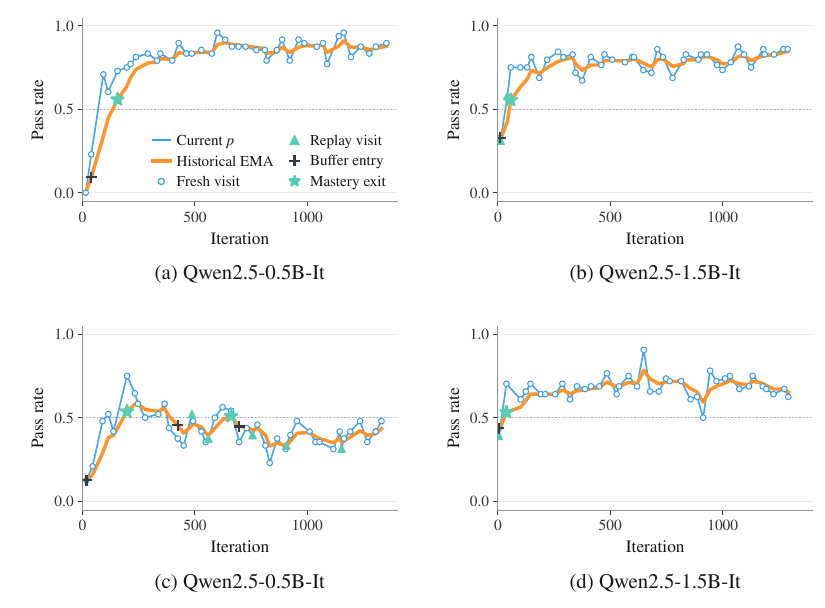}
    \captionsetup{skip=4pt}
    \caption{
    Population pass rates and historical EMA trajectories for four
    selected problems during DART-ES training on GSM8K.
    The top and bottom rows show improving and fluctuating cases,
    respectively.
    Blue and orange lines show the current population pass rate and
    the EMA updated with the current visit's feedback, respectively.
    Circles and triangles denote regular traversal and replay visits;
    plus signs mark buffer entry, and stars mark exit at the EMA
    mastery threshold of $0.5$, shown by the dotted line.
    Each case is selected closest to the median profile within its
    model and outcome group.
}
    \label{fig:difficulty_trajectories}
    \vspace{-10pt}
\end{figure}

The role of accumulated feedback is further reflected in state updates for individual problems and replay management. As shown in Figure~\ref{fig:difficulty_trajectories}, four cases from Qwen2.5-0.5B-It and Qwen2.5-1.5B-It belong to groups showing overall improvement or pronounced fluctuations. Within each model and group, we select problems whose profiles are close to the group median. The EMA in the figure includes feedback from the current visit, following sustained changes in pass rates while smoothing fluctuations in individual observations. Both improving cases exit the buffer when their historical EMA reaches the mastery threshold of $0.5$, while the fluctuating case for Qwen2.5-0.5B-It enters the buffer again after exiting. Since the mastery criterion depends on the historical EMA, a single pass rate above the threshold does not necessarily trigger an exit, and replay eligibility adjusts dynamically as feedback accumulates.

In summary, EMA improves predictions of subsequent pass rates in the middle and late stages of training and turns repeated population feedback into a continuously updated reference for each problem. Through this shared state, DART-ES connects perturbation direction evaluation with replay selection, allowing historical information to guide both the construction of update coefficients and the allocation of additional training opportunities. This highlights the design advantage of combining accumulated historical feedback with dynamic training decisions.

\subsection{Dynamic Maintenance and Visit Allocation in Targeted Replay}
\label{app:replay_dynamics}

DART-ES combines dynamic maintenance of the replay buffer with the allocation of additional visits, allowing replay to adapt to problem states and candidate availability. The buffer is continuously updated using historical pass rates and feedback from consecutive visits with no successful direction. Problems retained in the buffer must also satisfy the replay limit for each epoch and the cooldown interval before they can be sampled. By combining problem states with visit constraints, targeted replay continually adjusts which problems receive additional training opportunities and determines the actual number of replay visits according to the number of currently eligible candidates.

\begin{figure}[t]
    \centering
    \includegraphics[
        page=1,
        width=\linewidth
    ]{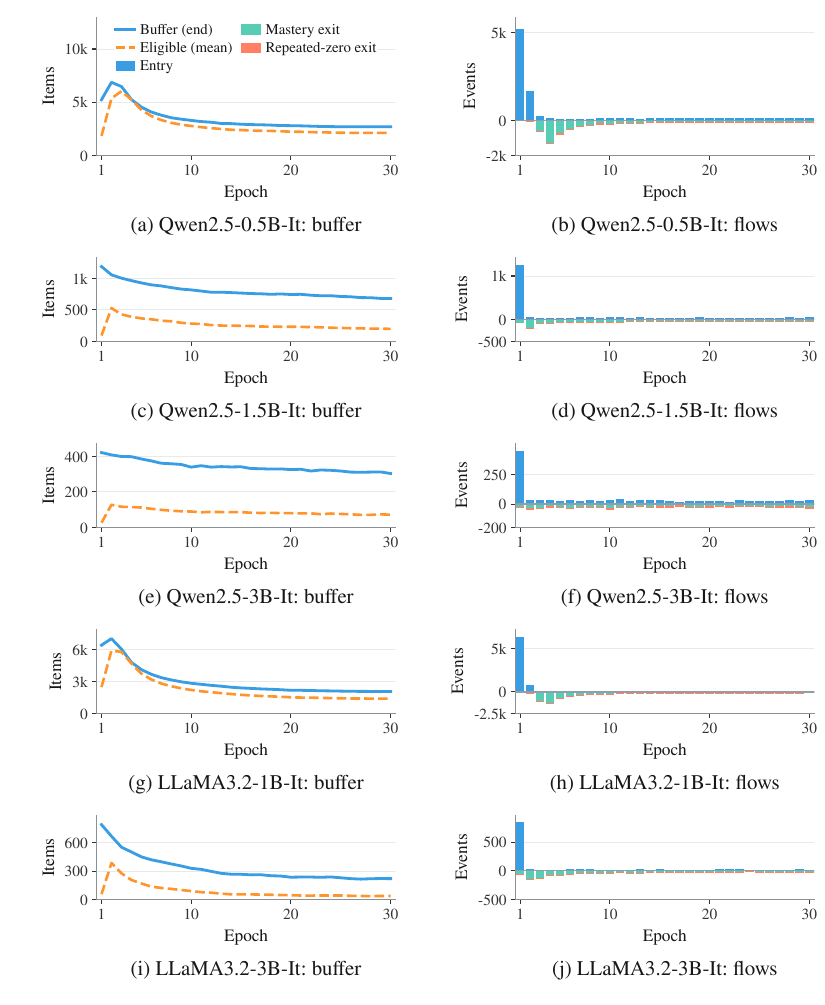}
    \captionsetup{skip=4pt}
    \caption{
    Replay buffer dynamics during DART-ES training on GSM8K.
    Each row corresponds to a base model.
    The left column shows buffer size at epoch end and the mean number
    of eligible candidates before sampling within each epoch.
    The right column shows entries above zero and exits below zero,
    distinguishing exits at the historical mastery threshold from
    exits after consecutive evaluations with no successful direction.
    A problem can enter and exit repeatedly.
    The legend in the first panel applies to all panels.
}
    \label{fig:replay_buffer_dynamics}
    \vspace{-10pt}
\end{figure}

This dynamic maintenance is reflected in continued changes in both buffer size and membership. As shown in Figure~\ref{fig:replay_buffer_dynamics}, the buffer reaches a relatively large size early in training and then generally shrinks, while problems continue to enter and exit. The plotted buffer size is measured at the end of each epoch, and the number of eligible candidates is averaged over the counts observed before sampling at each iteration within that epoch. For example, the buffer sizes for Qwen2.5-0.5B-It and LLaMA3.2-1B-It decrease from peaks of 6,867 and 7,012 to 2,683 and 2,061, respectively, at the end of the displayed period. Exits occur when the historical mastery threshold is reached or when the count of consecutive visits with no successful direction reaches the patience threshold. Across models, mastery exits account for 66.03\% to 90.97\% of all exit events within the displayed period. These two exit rules respond to historical mastery and the recent repeated absence of successful signals, respectively, allowing eligibility to remain in the buffer to change with feedback. Entries and exits are counted as events, and the same problem can enter and exit repeatedly. Even when the buffer size becomes relatively stable later in training, continued entry and exit events indicate that its membership remains dynamic, and the problems receiving additional visits change accordingly.

\begin{figure}[t]
    \centering
    \includegraphics[
        page=1,
        width=\linewidth
    ]{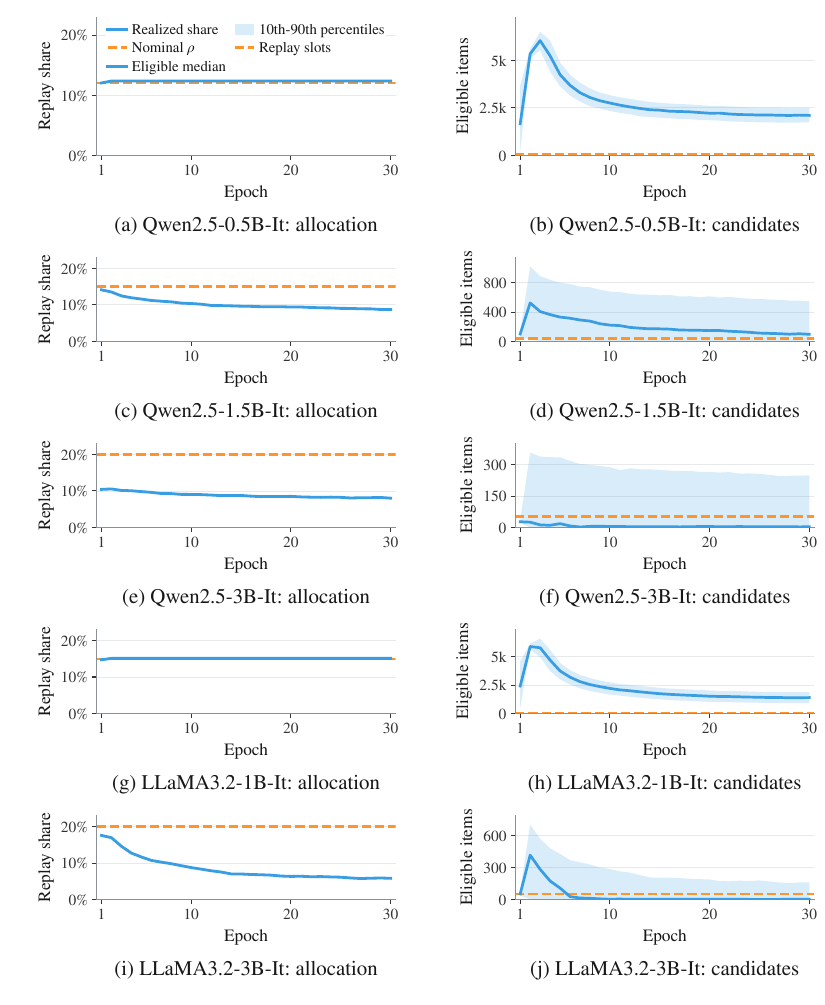}
    \captionsetup{skip=4pt}
    \caption{
    Realized replay allocation during DART-ES training on GSM8K.
    Each row corresponds to a base model.
    The left column compares the actual replay share in each epoch
    (total replay visits divided by total visits) with the nominal ratio $\rho$.
    The right column shows the median and the 10th to 90th percentiles
    of eligible candidate counts before sampling across iterations within each epoch.
    The dashed line on the right marks the planned replay slots
    $\operatorname{round}(B\rho)$, where $B$ is the nominal full batch size.
    The legend in the first panel applies to all panels.
}
    \label{fig:replay_allocation}
\end{figure}

These buffer updates, together with the availability of eligible candidates, determine how the planned replay allocation is carried out. As shown in Figure~\ref{fig:replay_allocation}, Qwen2.5-0.5B-It and LLaMA3.2-1B-It, for which candidates are generally sufficient, have realized replay ratios of 12.35\% and 15.10\% over the displayed period, close to their nominal ratios of 12\% and 15\%, respectively. The realized ratios for Qwen2.5-1.5B-It, Qwen2.5-3B-It, and LLaMA3.2-3B-It are 10.19\%, 8.88\%, and 8.57\%, below their respective nominal ratios of 15\%, 20\%, and 20\%. For these three models, iterations in which the number of candidates satisfying the current visit constraints falls below the number of planned replay slots account for 35.25\%, 62.35\%, and 63.46\% of all iterations in the displayed period, respectively. Analysis of individual iterations shows that the actual number of replay visits always equals the smaller of the number of planned slots and the number of currently eligible candidates. Thus, DART-ES follows the planned replay allocation when candidates are sufficient and limits additional visits to the available eligible candidates when their supply is limited. The actual allocation therefore respects both candidate eligibility and constraints on repeated visits. The realized replay ratio is computed as the total number of replay visits divided by the total number of visits. Slot rounding and smaller batches at the end of each epoch can also introduce small differences from the nominal ratio.

In summary, DART-ES continuously updates buffer membership so that the set of problems for replay changes with problem feedback, while candidate eligibility and visit constraints jointly regulate the number of additional visits. This design connects dynamic tracking of problem states with the actual allocation of training opportunities, allowing replay to adapt to changes in candidate states and availability while controlling repeated visits.

\clearpage

\section{Ablation Study}
\label{app:ablation_study}

\begin{table}[H]
    \centering
    \caption{
        Ablation results for DART-ES on GSM8K with
        Qwen2.5-0.5B-Instruct.
    }
    \label{tab:ablation_gsm8k}
    \begin{tabular}{@{}p{0.77\linewidth}r@{}}
        \toprule
        Method & Accuracy (\%) \\
        \midrule
        ES
            & 53.44 \\
        ES + continuous difficulty reweighting
            & 54.89 \\
        ES + targeted replay
            & 54.74 \\
        ES + historical difficulty estimation
        + continuous difficulty reweighting
            & 55.12 \\
        ES + historical difficulty estimation
        + targeted replay
            & 54.97 \\
        ES + continuous difficulty reweighting
        + targeted replay
            & 55.34 \\
        \rowcolor{dartgreen}
        DART-ES
            & 55.65 \\
        \bottomrule
    \end{tabular}
\end{table}

We conduct an ablation study on GSM8K with Qwen2.5-0.5B-Instruct
to examine the effects of historical difficulty estimation,
continuous difficulty reweighting, and targeted replay, as well as
their combinations. All configurations use the same base model,
training budget, and evaluation protocol. As shown in
Table~\ref{tab:ablation_gsm8k}, standard ES achieves an accuracy
of 53.44\%. Using reweighting or targeted replay alone increases
accuracy to 54.89\% and 54.74\%, respectively. These are gains of
1.45 and 1.30 percentage points. The results show that improving
perturbation direction evaluation and adjusting the allocation of
training visits can each make better use of problem level population
feedback. Reweighting affects direction evaluation in the current
population, while replay determines which problems receive
additional visits later in training. Using both mechanisms increases
accuracy to 55.34\%. This exceeds the configurations with only
reweighting or only replay by 0.45 and 0.60 percentage points,
respectively. The gain from their combination shows that these two
parts of training can work together. Feedback from replay visits
also enters direction evaluation through difficulty reweighting,
allowing the method to make better use of successful signals from
difficult problems.

Historical difficulty estimation further improves each of the three
matched configurations. With this mechanism, accuracy increases
from 54.89\% to 55.12\% for reweighting, from 54.74\% to 54.97\%
for replay, and from 55.34\% to 55.65\% when both are used.
The corresponding gains are 0.23, 0.23, and 0.31 percentage points.
These consistent improvements show that the historical difficulty
state provides a shared problem level reference for direction
evaluation and replay selection. In the combined
method, reweighting uses the historical state to adjust the effect
of successful directions on the current update. Targeted replay
uses the same state to select problems for later visits. Feedback
from those visits then updates the state. Together, the three
mechanisms form a coupled feedback loop. Problem level signals
from a population evaluation affect both the current update and
later visits, while the new feedback informs subsequent direction
evaluation and sample selection.

\clearpage

\section{Detailed Per-Step Runtime and GPU Memory Results}
\label{Detailed_Per-Step_Runtime_and_GPU_Memory_Results}















\textbf{Measurement Setup.}
We analyze the runtime and GPU memory usage of GRPO, GRPO+LoRA, and DART-ES. For each method and model, we use the corresponding training and generation configurations from the GSM8K experiments. GRPO and GRPO+LoRA are implemented using the VERL framework, with vLLM for rollout generation and Fully Sharded Data Parallel (FSDP) for distributed training. To reduce GPU memory usage, both methods enable gradient accumulation, padding removal, gradient checkpointing, activation offloading, actor parameter and optimizer state offloading, and reference model parameter offloading. DART-ES also uses vLLM for response generation during population evaluation but does not enable any of these additional memory-saving techniques. All experiments are conducted using four GPUs, each with 32 GiB of memory. Runtime per step is defined as the wall clock time required to complete one full training iteration. GPU memory usage is reported as the arithmetic mean of the peak
memory usage measured on each of the four GPUs. Figure~\ref{efficiency_comparison} provides a visual comparison of the results, while Tables~\ref{tab:runtime-five-models} and~\ref{tab:memory-five-models} report the complete numerical values.

\begin{table}[H]
\caption{Runtime per step for different methods across base models (s).}
\label{tab:runtime-five-models}
\begin{center}

\small
\setlength{\tabcolsep}{5pt}
\renewcommand{\arraystretch}{1.10}

\begin{tabular}{@{}lccccc@{}}

\hline
\noalign{\vskip 2pt}

\multicolumn{1}{c}{\bf Method} &
\multicolumn{3}{c}{\bf Qwen2.5-Instruct} &
\multicolumn{2}{c}{\bf Llama-3.2-Instruct}
\\

\cline{2-4}\cline{5-6}
\noalign{\vskip 1pt}

&
{\bf 0.5B} &
{\bf 1.5B} &
{\bf 3B} &
{\bf 1B} &
{\bf 3B}
\\

\noalign{\vskip 2pt}
\hline
\noalign{\vskip 2pt}

GRPO
& 76.67 
& 150.90
& 296.03
& 110.05
& 278.35 \\

GRPO+LoRA
& 146.50
& 251.76
& 419.10
& 160.75
& 666.92 \\

DART-ES
& 38.18
& 106.71
& 172.79
& 93.36
& 234.87 \\

\noalign{\vskip 2pt}
\hline

\end{tabular}

\end{center}
\vspace{-10pt}
\end{table}

\begin{table}[H]
\caption{Memory usage per GPU for different methods across base models (GiB).}
\label{tab:memory-five-models}
\begin{center}

\small
\setlength{\tabcolsep}{7pt}
\renewcommand{\arraystretch}{1.10}

\begin{tabular}{@{}lccccc@{}}

\hline
\noalign{\vskip 2pt}

\multicolumn{1}{c}{\bf Method} &
\multicolumn{3}{c}{\bf Qwen2.5-Instruct} &
\multicolumn{2}{c}{\bf Llama-3.2-Instruct}
\\

\cline{2-4}\cline{5-6}
\noalign{\vskip 1pt}

&
{\bf 0.5B} &
{\bf 1.5B} &
{\bf 3B} &
{\bf 1B} &
{\bf 3B}
\\

\noalign{\vskip 2pt}
\hline
\noalign{\vskip 2pt}

GRPO
& 9.71
& 15.01
& 22.61
& 13.52
& 22.50 \\

GRPO+LoRA
& 8.02
& 10.08
& 18.84
& 8.65
& 19.39 \\

DART-ES
& 4.75
& 8.56
& 12.81
& 8.04
& 17.75 \\

\noalign{\vskip 2pt}
\hline

\end{tabular}

\end{center}
\vspace{-10pt}
\end{table}

\textbf{Detailed Results.} As shown in Table~\ref{tab:runtime-five-models}, DART-ES achieves the shortest runtime per step on all five base models, and this advantage remains consistent across different model families and parameter scales. Within the Qwen2.5 family, the runtime per step of DART-ES is 38.18, 106.71, and 172.79 seconds on the 0.5B, 1.5B, and 3B models, respectively. These values are lower than the corresponding GRPO results of 76.67, 150.90, and 296.03 seconds, and are also clearly lower than the GRPO+LoRA results of 146.50, 251.76, and 419.10 seconds. On Llama3.2-1B and Llama3.2-3B, DART-ES requires 93.36 and 234.87 seconds per step, compared with 110.05 and 278.35 seconds for GRPO and 160.75 and 666.92 seconds for GRPO+LoRA. Overall, DART-ES reduces runtime per step by 15.2\%--50.2\% compared with GRPO and by 41.9\%--73.9\% compared with GRPO+LoRA.

Table~\ref{tab:memory-five-models} reports the peak memory usage per GPU under the same experimental settings. On Qwen2.5-0.5B, Qwen2.5-1.5B, and Qwen2.5-3B, DART-ES uses 4.75, 8.56, and 12.81 GiB, respectively. These values are lower than the GRPO results of 9.71, 15.01, and 22.61 GiB and the GRPO+LoRA results of 8.02, 10.08, and 18.84 GiB. On Llama3.2-1B and Llama3.2-3B, DART-ES uses 8.04 and 17.75 GiB, which are also lower than the corresponding GRPO results of 13.52 and 22.50 GiB and the GRPO+LoRA results of 8.65 and 19.39 GiB. Compared with GRPO, DART-ES reduces peak memory usage per GPU by 21.1\%--51.1\%. Compared with GRPO+LoRA, the reduction ranges from 7.1\% to 40.8\%. Runtime and memory usage generally increase with model size within each model family, but DART-ES consistently maintains the lowest runtime per step and peak memory usage per GPU at every evaluated scale. Notably, although DART-ES performs full parameter fine tuning, it still uses less GPU memory than the parameter efficient GRPO+LoRA baseline. This result shows that its efficiency gains do not come from reducing the number of trainable parameters, but are reflected in both execution speed and GPU memory requirements under the evaluated configurations.

\end{document}